\documentclass{article}
\usepackage[utf8]{inputenc}
\usepackage{microtype}
\usepackage{graphicx}
\usepackage{subcaption}
\usepackage{booktabs} %

\usepackage{hyperref}
\usepackage{xurl}

\usepackage[preprint]{icml2026}

\usepackage{amsmath}
\usepackage{amssymb}
\usepackage{mathtools}
\usepackage{amsthm}

\usepackage{booktabs}
\usepackage[table]{xcolor}

\usepackage{enumitem}

\usepackage[table]{xcolor}
\definecolor{yellow}{HTML}{f4e97a}

\usepackage[capitalize,noabbrev]{cleveref}

\usepackage{multirow}    %
\usepackage{graphicx}    %
\usepackage[table]{xcolor}  %

\theoremstyle{plain}

\theoremstyle{definition}

\theoremstyle{remark}

\usepackage[textsize=tiny]{todonotes}

\icmltitlerunning{Mashup Learning: Faster Finetuning by Remixing Past Checkpoints}
\makeatletter
\renewcommand{\@pa}[1]{%
  \ifcsname the@affil#1\endcsname
  \else
    \stepcounter{@affiliationcounter}%
    \newcounter{@affil#1}%
    \setcounter{@affil#1}{\value{@affiliationcounter}}%
  \fi
  \ifcsname @icmlsymbol#1\endcsname
    \expandafter\xdef\csname @icmlsymbolbynum\arabic{@affil#1}\endcsname{%
      \expandafter\noexpand\csname @icmlsymbol#1\endcsname}%
    \textsuperscript{\csname @icmlsymbol#1\endcsname\,}%
  \else
    \textsuperscript{\arabic{@affil#1}\,}%
  \fi
}

\renewcommand{\printAffiliationsAndNotice}[1]{\global\icml@noticeprintedtrue%
  \stepcounter{@affiliationcounter}%
  {\let\thefootnote\relax\footnotetext{\hspace*{-\footnotesep}\ificmlshowauthors #1\fi%
      \forloop{@affilnum}{1}{\value{@affilnum} < \value{@affiliationcounter}}{%
        \ifcsname @icmlsymbolbynum\arabic{@affilnum}\endcsname
          \textsuperscript{\csname @icmlsymbolbynum\arabic{@affilnum}\endcsname}%
        \else
          \textsuperscript{\arabic{@affilnum}}%
        \fi
        \ifcsname @affilname\the@affilnum\endcsname%
          \csname @affilname\the@affilnum\endcsname%
        \else
          {\bf AUTHORERR: Missing \textbackslash{}icmlaffiliation.}%
        \fi
      }.%
      \ifdefined\icmlcorrespondingauthor@text
         { }Correspondence to: \icmlcorrespondingauthor@text.%
      \else
        {\bf AUTHORERR: Missing \textbackslash{}icmlcorrespondingauthor.}%
      \fi
      \ \\
      \Notice@String
    }%
  }%
}
\makeatother

\begin{document}

\twocolumn[
  \icmltitle{Mashup Learning: Faster Finetuning by Remixing Past Checkpoints}

\icmlsetsymbol{aff1}{♪}

\begin{icmlauthorlist}
  \icmlauthor{Sofia Maria Lo Cicero Vaina}{aff1}
  \icmlauthor{Artem Chumachenko}{aff1}
  \icmlauthor{Max Ryabinin}{aff1}
\end{icmlauthorlist}
\icmlaffiliation{aff1}{Together AI}

\icmlcorrespondingauthor{Sofia Maria Lo Cicero Vaina}{s.lochichero@gmail.com}

  \icmlkeywords{Machine Learning, ICML}

  \vskip 0.3in
]

\printAffiliationsAndNotice{}  %

\begin{abstract}
Finetuning on domain-specific data is a well-established method for enhancing LLM performance on downstream tasks.
Training on each dataset produces a new set of model weights, resulting in a multitude of checkpoints saved in-house or on open-source platforms.
However, these training artifacts are rarely reused for subsequent experiments despite containing improved model abilities for potentially similar tasks.
In this paper, we propose Mashup Learning, a simple method to leverage the outputs of prior training runs to enhance model adaptation to new tasks.
Our procedure identifies the most relevant historical checkpoints for a target dataset, aggregates them with model merging, and uses the result as an improved initialization for training.
Across 8 standard LLM benchmarks, four models, and two collections of source checkpoints, Mashup Learning consistently improves average downstream accuracy by 0.5--5 percentage points over training from scratch.
It also accelerates convergence, requiring 41--46\% fewer training steps and up to 37\% less total wall-clock time to match from-scratch accuracy, including all selection and merging overhead.
\end{abstract}

\section{Introduction}

The training process for foundation models could be split into two stages: pretraining on vast amounts of general-domain data, followed by post-training on smaller datasets to specialize for particular tasks~\cite{llama3,qwen3,olmo2025olmo3}.
Given that the pretraining stage is more compute-intensive, the base model obtained after this stage is often finetuned multiple times on different datasets and with varied hyperparameter combinations.
While this procedure is relatively inexpensive, its computational cost can become more significant over multiple iterations to find the best training setup for a general-purpose model~\cite{modelsoup,cohere2025commandaenterprisereadylarge} or to adapt the base checkpoint to multiple downstream scenarios~\cite{wei2022finetunedlanguagemodelszeroshot,wang-etal-2022-super}.
Moreover, training the largest models even on small datasets can present a challenge for academic researchers and enthusiasts due to the associated engineering and hardware requirements~\cite{qlora, lv-etal-2024-full}.
Lastly, achieving the best results on a given task through finetuning might be complicated due to a limited number of training examples available in the provided dataset.

\begin{figure}
    \centering
    \includegraphics[width=\linewidth]{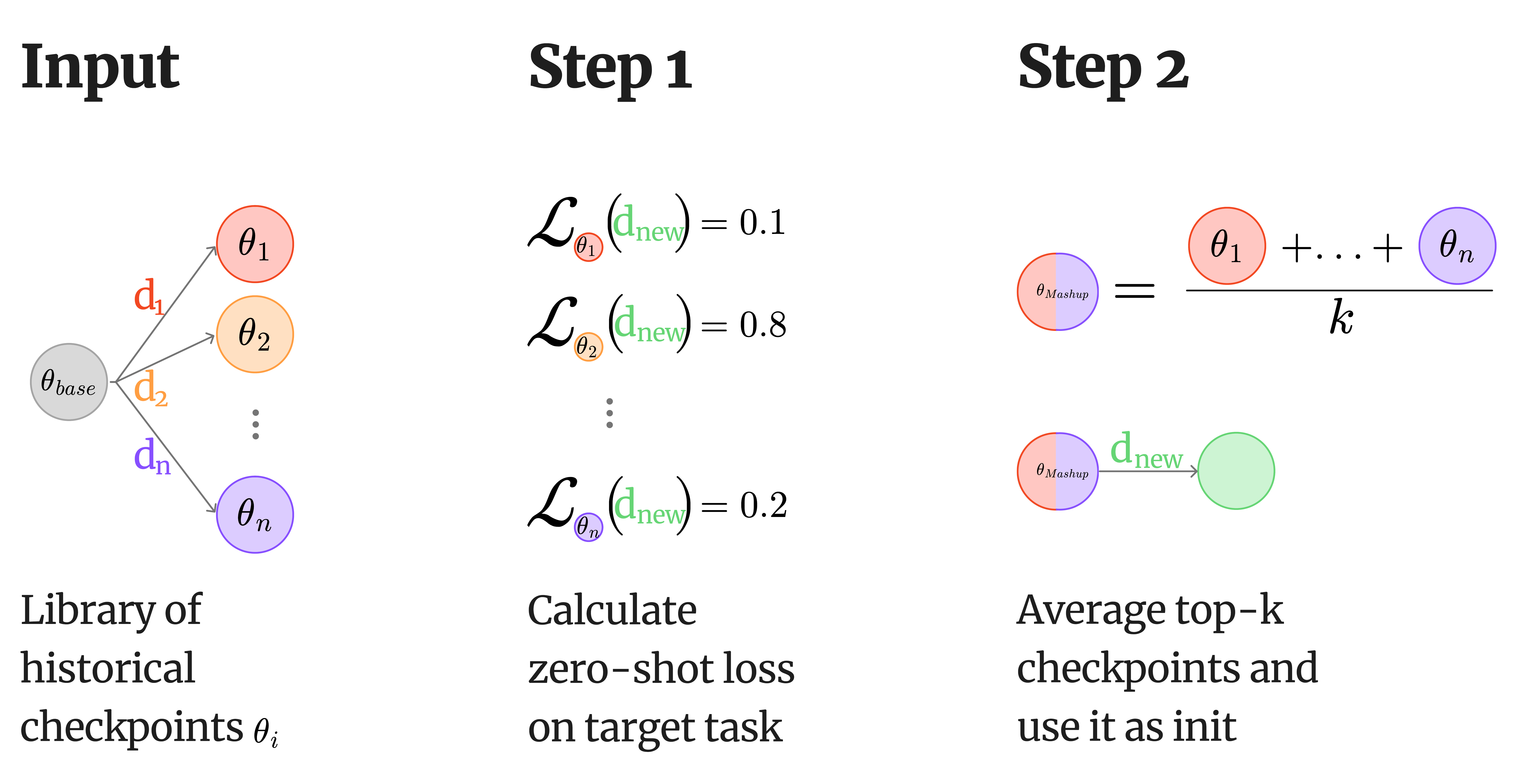}
    \caption{A schematic example of Mashup Learning.}
    \label{fig:scheme}
\end{figure}

On the other hand, running multiple finetuning experiments generates a rich collection of checkpoints that can serve as historic data --- whether from continual learning, resumed training, hyperparameter sweeps, or the thousands of community-published adapted versions of open models\footnote{For example, there are \href{https://huggingface.co/models?other=base_model:finetune:meta-llama/Llama-3.1-8B-Instruct}{over 2,000 finetuned versions} of Llama 3.1-8B-Instruct on the Hugging Face Hub.}.
Prior work has shown that such checkpoint collections can be combined to improve model quality without training, e.g.\ by averaging models from different hyperparameter runs~\cite{modelsoup} or by merging with earlier checkpoints during training to mitigate catastrophic forgetting~\cite{souptogo}.
This, together with the general success of transfer learning~\cite{le2012buildinghighlevelfeaturesusing,howard-ruder-2018-universal}, leads us to hypothesize that model adaptation might benefit from other finetuned checkpoints as well.
However, to the best of our knowledge, no prior work has explored using merged checkpoints as an initialization for finetuning on new tasks.
Furthermore, recent research demonstrates that the parameter space of trained versions for the same model has a low-dimensional structure, which can be leveraged for model reuse~\cite{kaushik2025universalweightsubspacehypothesis}. 
Another branch of research dedicated to data mixtures finds evidence that different tasks can be characterized through linear combinations of meta-tasks \cite{domain2vec,doremi}. 
This suggests that models get repeatedly trained on the same meta-tasks, which creates an opportunity for optimization. 

Inspired by the above observations, we suggest taking a different approach to model post-training, which \textit{leverages the information} from previous finetuned checkpoints and does not discard the compute resources that were spent on their corresponding experiments.
We propose a new model adaptation paradigm, named \textbf{Mashup Learning}, which recycles model checkpoints trained on \textit{prior} datasets to provide a set of better initial weights for a new task.
The checkpoints could be chosen from a collection of arbitrary size based on a criterion that measures the training loss for each of those checkpoints on a small subsample of the new dataset.
Mashup Learning is easy to implement, does not require any changes to the training procedure itself, and could be scaled to large checkpoint collections due to the embarrassingly parallel nature of relevance estimation.
In practice, Mashup Learning produces stronger models when training with the same token budget, and reaches the same quality as training from scratch in fewer iterations.
An illustration of Mashup Learning is displayed in Figure~\ref{fig:scheme}.

Our contributions are as follows:
\begin{enumerate}[itemsep=0.5em, topsep=0.5em, parsep=0.5em]
    \item We propose Mashup Learning, a model- and domain-agnostic method that leverages historical checkpoints to construct custom initializations for finetuning on new tasks, without requiring any modifications to the training procedure. 
    To the best of our knowledge, this is the first method to repurpose historical checkpoints for improved finetuning initialization.
    \item We evaluate Mashup Learning in its simplest form: selecting checkpoints by loss on the target task and averaging them to obtain an initialization. This procedure yields consistent improvements across Gemma-3 4B, Gemma-3 1B, Gemma-2 2B, and Mistral-7B-Instruct-v0.2 on 8 standard LLM benchmarks with two collections of source checkpoints\footnote{The code to reproduce our experiments is publicly available at \href{https://github.com/2son1a/mashup-learning}{\texttt{github.com/2son1a/mashup-learning}}.}. Across all configurations, Mashup Learning improves average downstream accuracy by 0.5--5 percentage points while accelerating convergence, matching from-scratch accuracy in 41--46\% fewer steps and up to 37\% less wall-clock time, including all overhead.
    \item We thoroughly verify our design choices and show that performance can be further improved by replacing averaging with model merging techniques, and by selecting checkpoints through task-specific metrics.
\end{enumerate}

\section{Background}
Modern large language models~\cite{llama3, qwen3, kimik2} are capable of solving general problems.
Still, a broad number of tasks benefit from further training on domain-specific data, which is called finetuning. 
During finetuning, practitioners aim to achieve the best quality on their validation dataset while spending less money on collecting custom datasets, incurring less storage overhead, and spending fewer GPU hours on re-training models in search of optimal hyperparameters~\cite{lorawithoutregret}. 
These challenges motivate multiple methods that improve data efficiency, parameter efficiency, and hyperparameter robustness, respectively~\cite{peft, ether, dataefficientsft}.

In our experiments, along with full finetuning we train Low-Rank Adaptation (LoRA,~\citealp{lora}) adapters. 
LoRA is a parameter-efficient finetuning method~\cite{peft} that freezes model weights and injects low-rank trainable matrices, reducing the number of trainable parameters and the memory footprint of trained checkpoints. 
LoRA has been shown to match the performance of full finetuning on smaller tasks~\cite{loralearns}.

\section{Method}

In this section, we describe the Mashup Learning method for initializing models using historical checkpoints before finetuning on a new downstream task.

Mashup Learning requires a library of checkpoints trained on various downstream tasks. 
These checkpoints must share the same architecture as the target model. 
For LoRA-based models, hyperparameters such as target modules, rank, and $\alpha$ must also match. 
Such checkpoints can be obtained from open-source repositories like Hugging Face Hub or collected in-house.
In the latter case, the initial parameter values of historical experiments can also be collected, enabling more sophisticated model merging methods. 
We elaborate on this topic in Section~\ref{subsec:mergingmethods}.

The next step is to identify the checkpoints from the library that will be merged. 
We evaluate each checkpoint on a small subset of the target task's training data and select the top-$k$ checkpoints with the lowest loss value.
We use the training set to ensure that the method is applicable in practice, when validation data is not available.
Our analysis in Section~\ref{subsec:eval_size} demonstrates that 256 samples provide sufficient signal for checkpoint selection, with larger evaluation sets yielding diminishing returns.
Furthermore, Section~\ref{subsec:checkpoint_selection} shows that this selection method efficiently approximates the ``oracle'' performance of choosing the best possible combination. 
When available, task-specific metrics such as accuracy can be used instead of perplexity to further improve final performance.

Finally, we aggregate the selected checkpoints, obtaining a single set of model parameters, and use the result as initialization for training on the target task.
While simple averaging can be used, more advanced model merging methods may yield better results by resolving conflicting parameters across checkpoints.
Some of these methods require the model initialization in order for task vectors/delta parameters to be computed. For example, DARE-TIES we found to be top-performing in our experiments.
In the case of full finetuning, it is natural to expect that the initial foundation model weights are available for the given historical checkpoints.
However, in the case of LoRAs, initializations are typically not present.
We compare the performance of different merging methods and explore the optimal number of checkpoints to select in Section~\ref{subsec:mergingmethods}.

The complete description of Mashup Learning is outlined in Algorithm~\ref{alg:mashup}.

\begin{algorithm}[t]
\caption{Mashup Learning}
\label{alg:mashup}
\begin{algorithmic}[1]
\REQUIRE Target task dataset $\mathcal{D} = (\mathcal{D}_\mathrm{train}, \mathcal{D}_\mathrm{val})$
\REQUIRE Set of $N$ model checkpoints $\{\theta_1, \ldots, \theta_N\}$ with matching architecture
\ENSURE Initialized parameters $\theta^*$ for finetuning
\STATE \textbf{Step 1: Rank checkpoints by task performance}
\FOR{$i = 1, \ldots, N$}
    \STATE $\ell_i \gets \mathcal{L}(\theta_i; \mathcal{D}_\mathrm{train})$ \COMMENT{Evaluate loss}
\ENDFOR
\STATE $\pi \gets \mathrm{argsort}([\ell_1, \ldots, \ell_N])$ \COMMENT{Sort ascending}
\STATE \textbf{Step 2: Average top-$k$ checkpoints}
\STATE $\theta^* \gets \frac{1}{k} \sum_{i=1}^{k} \theta_{\pi_i}$
\STATE \textbf{Step 3: Fine-tune on the target task}
\STATE $\theta_\mathrm{final} \gets \mathrm{Train}(\theta^*; \mathcal{D}_\mathrm{train})$
\STATE \textbf{Return} $\theta_\mathrm{final}$
\end{algorithmic}
\end{algorithm}

\section{Experiments}

In this section, we summarize the evaluation of Mashup Learning on finetuning large language models. 
Our primary goal is to study the effectiveness of the method and the extent of its advantages compared to training from scratch.

\subsection{Setup}

We conduct all experiments with large language models using the Transformer~\cite{transformer} architecture.
We focus on this class of models due to the broad availability of both downstream NLP benchmarks and pretrained models that benefit from training on such benchmarks.
This choice also aligns with our related work, where the merging and adaptation methods we compare against were designed for and evaluated on Transformer models.
Although it would be highly valuable to explore other modalities and architectures, here we aim to focus on the behavior of the method itself and leave such an exploration to future work.

\paragraph{Datasets} 
We evaluate all methods on a variety of academic datasets commonly used to benchmark modern foundation models~\cite{gemma3}.
Additionally, these datasets represent popular downstream tasks for finetuning and are frequently used in our related work~\cite{text2lora,dare}.
We use the following eight benchmarks: ARC-Easy~\cite{clark2018think}, which evaluates grade-school reasoning; OpenBookQA~\cite{mihaylov2018openbookqa}, which contains open-book question answering problems; WinoGrande~\cite{sakaguchi2019winogrande} and PIQA~\cite{bisk2020piqa}, which assess commonsense and physical reasoning; MathQA~\cite{amini2019mathqa}, which tests mathematical reasoning; HellaSwag~\cite{zellers2019hellaswag}, which measures commonsense natural language inference; SocialIQA~\cite{sap2019socialiqa}, which benchmarks social interaction understanding; and CommonsenseQA~\cite{talmor2019commonsenseqa}, which evaluates commonsense question answering.

\begin{table}[t]
    \centering
    \caption{Dataset summary for our primary experiments.}
    \setlength{\tabcolsep}{3pt}
    \begin{tabular}{llcc}
    \toprule
    \textbf{Dataset} & \textbf{Short name} & \textbf{Train size} & \textbf{Val size} \\
    \midrule
    ARC-Easy & ARC-e & 2,251 & 570 \\
    CommonsenseQA & CSQA & 9,741 & 1,221 \\
    HellaSwag & Hella. & 39,905 & 10,042 \\
    MathQA & MathQA & 29,837 & 4,475 \\
    OpenBookQA & OBQA & 4,957 & 500 \\
    PIQA & PIQA & 16,113 & 1,838 \\
    SocialIQA & SIQA & 33,410 & 1,954 \\
    WinoGrande & Wino. & 40,398 & 1,267 \\
    \bottomrule
    \end{tabular}
    \label{tab:dataset_summary}
\end{table}

A summary of dataset statistics is given in Table~\ref{tab:dataset_summary}. 
Every benchmark is a multiple-choice task, a model is required to select the correct answer.
To conduct the evaluation, we provide the model with the possible answers, represented as their corresponding tokens. 
Then, we prompt the model to generate a token for the answer, counting the example as solved if the answer matches the ground truth, and assess the quality of its response by measuring accuracy on the validation set. 
The evaluation prompts can be found in our GitHub repository.

\paragraph{Experiment protocol}

Mashup Learning requires a collection of checkpoints for task-specific model initialization.
In our main experiments, we use the eight datasets described above in a leave-one-out setup: each dataset is treated in turn as the target, while models trained on the remaining seven serve as checkpoint sources.
Training from scratch on each dataset provides the baselines against which we evaluate Mashup Learning. To verify the generality of our approach, we conduct experiments in two popular setups: full-parameter finetuning and LoRA training~\cite{lora}.

\begin{table*}[t]
    \centering
    \caption{\textbf{Finetuning methods comparison on Mistral-7B-Instruct-v0.2, Lots-of-LoRAs setup.} Accuracy results across six benchmarks comparing different LoRA merging and initialization approaches. The best result is highlighted in bold.} 
    \label{tab:compare_baselines}
    \begin{tabular}{lccccccc}
        \toprule
        \textbf{Method} & \textbf{ARC-e} & \textbf{Hella.} & \textbf{MathQA} & \textbf{OBQA} & \textbf{PIQA} & \textbf{Wino.} & \textbf{Avg.} \\
        \midrule
        From scratch & 84.0 & 93.2 & 21.7 & 74.6 & 85.2 & 81.4 & 73.4 \\
        \midrule
        Text-to-LoRA & 82.4 & 47.6 & 26.3 & 71.2 & 80.2 & 66.1 & 62.3 \\
        Text-to-LoRA used as init & 83.1 & 91.6 & 26.2 & 74.6 & 85.1 & 74.2 & 72.5 \\
        \midrule
        Mashup Learning at init & 85.0 & 61.2 & 26.4 & 77.9 & 80.6 & 64.9 & 66.0 \\
        Mashup Learning  & \textbf{87.9} & \textbf{95.1} & \textbf{32.1} & \textbf{86.6} & \textbf{86.5} & \textbf{82.5} & \textbf{78.5} \\
        \bottomrule
    \end{tabular}
\end{table*}

\paragraph{Models}
We use Gemma-3 4B, Gemma-3 1B, and Gemma-2 2B~\cite{gemma2, gemma3} as our primary models, as they are the top three models with fewer than 5B parameters on the LM Arena leaderboard~\cite{chiang2024chatbot}, performing on par with larger models such as Ministral-8B and Llama-3-8B-Instruct; the full ranking is provided in Table~\ref{tab:arena_small_models}. To compare against~\citet{text2lora}, which relies on pretrained checkpoints from~\citet{brüelgabrielsson2025compressserveservingthousands}, we additionally experiment with Mistral-7B-Instruct-v0.2~\cite{jiang2023mistral7b}.

\paragraph{Baselines} 
In the case of LoRA training, we compare Mashup Learning with two primary baselines: random initialization of parameters~\cite{lora} and the Text-to-LoRA model~\cite{text2lora}.
For random initialization, we use a random Gaussian initialization for $A$ and zeros for $B$.
For Text-to-LoRA, we use the model provided by the authors to generate adapters for each task. 
The hypernetwork requires a text description of a task as input. 
We use the prompts provided by the authors of the original paper.
Each task has three different prompts, and we report their aggregated metrics in Table~\ref{tab:compare_baselines}.

\paragraph{Hyperparameters}
Learning rates for full finetuning and LoRA are swept over $[5\mathrm{e}{-6}, 5\mathrm{e}{-5}]$ and $[5\mathrm{e}{-5}, 5\mathrm{e}{-4}]$, respectively, and tuned separately for each model and dataset, following~\citet{lee2026learningratemattersvanilla}.
In experiments where we use LoRA, we target all Transformer modules with a rank of 8 and $\alpha = 2r$, following the best practices proposed by~\citet{loralearns}.
One of our experiments includes a comparison with a Text-to-LoRA baseline that recycles the Lots-of-LoRAs collection; for this experiment, LoRA hyperparameters are adjusted to be comparable with theirs.
Full hyperparameters for both full finetuning and LoRA are reported in Appendix~\ref{app:hyperparams}.

\paragraph{Computational resources}Individual LoRA runs average 5--8 minutes and full finetuning runs 15--18 minutes on a single H100 GPU. 
The complete set of experiments required approximately 500 GPU-hours on H100 GPUs.

\subsection{Results}

\begin{table*}[t]
    \centering
    \caption{\textbf{Mashup Learning vs.\ training from scratch, leave-one-out setup.} Average accuracy results across eight benchmarks over three seeds. The best result for each model is highlighted in bold.}
    \label{tab:combined_results_all}
    \resizebox{\linewidth}{!}{
    \begin{tabular}{lllccccccccc}
        \toprule
        \textbf{Model} & \textbf{Setup} & \textbf{Method} & \textbf{ARC-e} & \textbf{CSQA} & \textbf{Hella.} & \textbf{MathQA} & \textbf{OBQA} & \textbf{PIQA} & \textbf{SIQA} & \textbf{Wino.} & \textbf{Avg.} \\
        \midrule
        \multirow{4}{*}{Gemma-3 1B} & \multirow{2}{*}{LoRA} & From scratch & 75.0 & 71.9 & 81.4 & 39.3 & 70.3 & 77.0 & \textbf{74.2} & 70.6 & 70.0 \\
        & & Mashup Learning & \textbf{78.8} & \textbf{73.5} & \textbf{82.7} & \textbf{39.8} & \textbf{75.6} & \textbf{78.4} & 74.1 & \textbf{71.2} & \textbf{71.8} \\
        \cmidrule{2-12}
         & \multirow{2}{*}{Full FT} & From scratch & 73.9 & 68.8 & 79.8 & 37.3 & 68.3 & 76.1 & 73.0 & 69.2 & 68.3 \\
        & & Mashup Learning & \textbf{78.1} & \textbf{70.5} & \textbf{80.2} & \textbf{39.1} & \textbf{73.1} & \textbf{77.7} & \textbf{73.3} & \textbf{69.6} & \textbf{70.2} \\
        \midrule
        \multirow{4}{*}{Gemma-2 2B} & \multirow{2}{*}{LoRA} & From scratch & 90.1 & 82.5 & 93.5 & 44.7 & 85.0 & 85.3 & 81.5 & 83.6 & 80.8 \\
        & & Mashup Learning & \textbf{91.6} & \textbf{82.7} & \textbf{93.6} & \textbf{45.8} & \textbf{85.5} & \textbf{86.3} & \textbf{81.7} & \textbf{84.4} & \textbf{81.5} \\
        \cmidrule{2-12}
         & \multirow{2}{*}{Full FT} & From scratch & 89.3 & 77.7 & 91.1 & 40.5 & 80.8 & 83.0 & 79.0 & 78.6 & 77.5 \\
        & & Mashup Learning & \textbf{90.1} & \textbf{78.4} & \textbf{91.7} & \textbf{40.7} & \textbf{81.2} & \textbf{83.1} & \textbf{79.5} & \textbf{78.9} & \textbf{78.0} \\
        \midrule
        \multirow{4}{*}{Gemma-3 4B} & \multirow{2}{*}{LoRA} & From scratch & 91.6 & 83.2 & 94.3 & 54.9 & 85.6 & 87.9 & \textbf{82.4} & 85.9 & 83.2 \\
        & & Mashup Learning & \textbf{93.5} & \textbf{83.8} & \textbf{94.7} & \textbf{56.9} & \textbf{87.7} & \textbf{88.5} & 82.2 & \textbf{86.2} & \textbf{84.2} \\
        \cmidrule{2-12}
         & \multirow{2}{*}{Full FT} & From scratch & 90.7 & 80.6 & 92.6 & 46.2 & 83.9 & \textbf{86.1} & 79.5 & 81.9 & 80.2 \\
        & & Mashup Learning & \textbf{92.3} & \textbf{80.7} & \textbf{92.7} & \textbf{46.5} & \textbf{84.6} & 85.9 & \textbf{81.0} & \textbf{83.3} & \textbf{80.9} \\
        \bottomrule
    \end{tabular}
    }
\end{table*}

\paragraph{Accuracy}
The results of our primary experiments in the leave-one-out setup are presented in Table~\ref{tab:combined_results_all}.
Across all three model families and both training regimes, Mashup Learning consistently improves over training from scratch.
For LoRA finetuning, average accuracy increases by 1.8, 0.7, and 1.0 percentage points on Gemma-3 1B, Gemma-2 2B, and Gemma-3 4B, respectively.
Full finetuning exhibits comparable gains: +1.9 on Gemma-3 1B and +0.5 on Gemma-2 2B.
The improvements are broad rather than concentrated on a single task---Mashup Learning matches or exceeds the from-scratch baseline on nearly every benchmark, with the largest per-task gains observed on OpenBookQA (+5.3 for Gemma-3 1B LoRA) and ARC-Easy (+4.2 for Gemma-3 1B full FT).

\paragraph{Convergence speed}
Beyond final accuracy, Mashup Learning accelerates convergence (Table~\ref{tab:combined_speedup_all}).
Even from-scratch training often reaches 99\% of its own converged accuracy well before training ends (at 59--79\% of steps on average), indicating that later steps yield diminishing returns.
Mashup Learning converges substantially faster still: on average, it matches the converged from-scratch accuracy after only 51--59\% of training, depending on the model and setup.
LoRA benefits the most, with Mashup converging at 55.5--59.1\% of training versus 69.4--79.4\% from scratch.
For full finetuning, Mashup reaches parity at 51.8--56.1\% compared to 59.0--67.3\% from scratch.
Individual tasks can converge much earlier (e.g., ARC-Easy at just 9.0\% for Gemma-3 4B LoRA).

\paragraph{Wall-clock time}

Table~\ref{tab:combined_timing_all} reports end-to-end wall-clock time, including the overhead of relevance estimation and checkpoint merging.
Even with this overhead, Mashup Learning reduces total training time in the large majority of configurations, with full finetuning completing in 63--81\% of the from-scratch wall-clock time on average and LoRA in 86--88\%.
In a small number of LoRA cases (e.g., SIQA on Gemma-3 1B and Gemma-3 4B), the overhead slightly exceeds the convergence savings, resulting in ratios marginally above 100\%.

\paragraph{Comparison with baselines}

Text-to-LoRA~\cite{text2lora} showed improved results over model merging in zero-shot adaptation.
To compare, we fine-tuned the adapters generated by Text-to-LoRA on the target dataset and report both the original and fine-tuned results.
Since their method uses Lots-of-LoRAs collection of 1,000 adapters trained on a variety of tasks, we used this same collection as a source of checkpoints for Mashup Learning.

As shown in Table~\ref{tab:compare_baselines}, Mashup Learning achieves the highest results on Mistral-7B-Instruct-v0.2, outperforming all alternatives by a wide margin (+5.1 average accuracy over training from scratch).
Both the merged initialization without further training (Mashup Learning at init) and the Text-to-LoRA-generated adapters show improvements on individual tasks, but only Mashup Learning with continued fine-tuning yields consistent gains across all benchmarks.

Note that this experiment uses six benchmarks rather than the full set used elsewhere, as these are the ones for which the original paper provided prompts.

\begin{table*}[t]
    \centering
    \caption{\textbf{Convergence speedup across all models.} Percentage of training steps at which each method reached 99\% of converged from-scratch accuracy. Lower is better. ``---'' if any seed did not reach 99\% of from-scratch accuracy.}
    \label{tab:combined_speedup_all}
    \resizebox{\linewidth}{!}{
    \begin{tabular}{lllccccccccc}
    \toprule
    \textbf{Model} & \textbf{Setup} & \textbf{Method} & \textbf{ARC-e} & \textbf{CSQA} & \textbf{Hella.} & \textbf{MathQA} & \textbf{OBQA} & \textbf{PIQA} & \textbf{SIQA} & \textbf{Wino.} & \textbf{Avg.} \\
    \midrule
    \multirow{4}{*}{Gemma-3 1B} & \multirow{2}{*}{LoRA} & From scratch & 74.1 & 80.3 & 73.9 & \textbf{88.8} & 82.7 & 73.0 & 78.5 & 84.1 & 79.4 \\
     & & Mashup & \textbf{28.9} & \textbf{48.5} & \textbf{61.1} & --- & \textbf{43.1} & \textbf{47.4} & \textbf{75.0} & \textbf{69.1} & \textbf{59.1} \\
    \cmidrule{2-12}
     & \multirow{2}{*}{Full FT} & From scratch & 62.8 & 65.5 & 67.8 & 72.4 & 74.2 & 60.3 & 70.8 & \textbf{64.8} & 67.3 \\
     & & Mashup & \textbf{9.6} & \textbf{53.5} & \textbf{61.0} & \textbf{55.5} & \textbf{41.0} & \textbf{42.1} & \textbf{51.4} & --- & \textbf{51.8} \\
    \midrule
    \multirow{4}{*}{Gemma-2 2B} & \multirow{2}{*}{LoRA} & From scratch & 53.7 & 72.2 & 65.0 & 80.8 & 79.7 & 66.1 & \textbf{71.2} & 72.5 & 70.2 \\
     & & Mashup & \textbf{11.7} & \textbf{69.4} & \textbf{58.1} & \textbf{74.8} & \textbf{64.9} & \textbf{55.8} & 73.5 & \textbf{64.0} & \textbf{59.0} \\
    \cmidrule{2-12}
     & \multirow{2}{*}{Full FT} & From scratch & --- & 62.3 & 47.5 & 64.1 & \textbf{61.7} & 39.1 & 44.4 & \textbf{52.8} & 59.0 \\
     & & Mashup & \textbf{0} & \textbf{58.3} & \textbf{40.8} & \textbf{60.8} & --- & \textbf{30.3} & \textbf{39.7} & --- & \textbf{53.7} \\
    \midrule
    \multirow{4}{*}{Gemma-3 4B} & \multirow{2}{*}{LoRA} & From scratch & 43.1 & 71.1 & 62.2 & 81.6 & 69.6 & 74.8 & 74.4 & 78.6 & 69.4 \\
     & & Mashup & \textbf{9.0} & \textbf{64.9} & \textbf{53.2} & \textbf{67.2} & \textbf{46.5} & \textbf{53.3} & \textbf{71.9} & \textbf{78.0} & \textbf{55.5} \\
    \cmidrule{2-12}
     & \multirow{2}{*}{Full FT} & From scratch & 32.3 & \textbf{64.9} & 65.9 & \textbf{84.1} & \textbf{62.7} & 68.1 & 41.6 & 67.6 & 60.9 \\
     & & Mashup & \textbf{11.7} & 69.3 & \textbf{64.0} & --- & 68.3 & \textbf{50.7} & \textbf{26.9} & \textbf{58.1} & \textbf{56.1} \\
    \bottomrule
\end{tabular}
    }
\end{table*}

\begin{table*}[t]
    \centering
    \caption{\textbf{Wall-clock training time across all models.} Training time in seconds at the best learning rate. Mashup includes relevance estimation and merge overhead, with training time scaled by convergence percentage. Ratio $<100\%$ means Mashup is faster.}
    \label{tab:combined_timing_all}
    \small
    \setlength{\tabcolsep}{5pt}
    \begin{tabular}{lllccccccccc}
        \toprule
        \textbf{Model} & \textbf{Setup} & & \textbf{ARC-e} & \textbf{CSQA} & \textbf{Hella.} & \textbf{MathQA} & \textbf{OBQA} & \textbf{PIQA} & \textbf{SIQA} & \textbf{Wino.} & \textbf{Avg.} \\
        \midrule
        \multirow{6}{*}{Gemma-3 1B} & \multirow{3}{*}{LoRA} & From scratch (s) & 101 & 162 & 611 & 356 & 134 & 261 & 365 & 400 & 299 \\
        & & Mashup (s) & 90 & 146 & 438 & 337 & 99 & 181 & 400 & 356 & 256 \\
        & & Ratio (\%) & 89\% & 90\% & 72\% & 95\% & 73\% & 69\% & 109\% & 89\% & 86\% \\
        \cmidrule{2-12}
         & \multirow{3}{*}{Full FT} & From scratch (s) & 298 & 308 & 2799 & 1801 & 231 & 842 & 1026 & 1021 & 1041 \\
        & & Mashup (s) & 127 & 188 & 2001 & 1177 & 136 & 398 & 656 & 991 & 709 \\
        & & Ratio (\%) & 42\% & 61\% & 72\% & 65\% & 59\% & 47\% & 64\% & 97\% & 63\% \\
        \midrule
        \multirow{6}{*}{Gemma-2 2B} & \multirow{3}{*}{LoRA} & From scratch (s) & 152 & 237 & 1120 & 639 & 188 & 454 & 555 & 664 & 501 \\
        & & Mashup (s) & 81 & 258 & 980 & 565 & 193 & 346 & 547 & 505 & 434 \\
        & & Ratio (\%) & 53\% & 109\% & 88\% & 88\% & 103\% & 76\% & 99\% & 76\% & 87\% \\
        \cmidrule{2-12}
         & \multirow{3}{*}{Full FT} & From scratch (s) & 307 & 404 & 2151 & 1297 & 288 & 811 & 1038 & 1066 & 920 \\
        & & Mashup (s) & 133 & 361 & 1235 & 1073 & 238 & 688 & 720 & 806 & 657 \\
        & & Ratio (\%) & 44\% & 90\% & 57\% & 83\% & 83\% & 86\% & 69\% & 75\% & 73\% \\
        \midrule
        \multirow{6}{*}{Gemma-3 4B} & \multirow{3}{*}{LoRA} & From scratch (s) & 96 & 186 & 1170 & 600 & 136 & 430 & 502 & 539 & 457 \\
        & & Mashup (s) & 54 & 197 & 974 & 452 & 123 & 376 & 547 & 518 & 405 \\
        & & Ratio (\%) & 56\% & 106\% & 83\% & 75\% & 90\% & 87\% & 109\% & 96\% & 88\% \\
        \cmidrule{2-12}
         & \multirow{3}{*}{Full FT} & From scratch (s) & 289 & 445 & 2595 & 1593 & 349 & 753 & 1174 & 1462 & 1083 \\
        & & Mashup (s) & 126 & 265 & 2639 & 1640 & 414 & 800 & 430 & 1135 & 931 \\
        & & Ratio (\%) & 43\% & 60\% & 102\% & 103\% & 119\% & 106\% & 37\% & 78\% & 81\% \\
        \bottomrule
    \end{tabular}
\end{table*}

\section{Analysis}

In this section, we study the design decisions that determine the exact form of Mashup Learning, aiming to understand which factors play a critical role in the performance of the method. We examine how to select relevant checkpoints from a library (Section~\ref{subsec:checkpoint_selection}), how much data is needed for this selection (Section~\ref{subsec:eval_size}), which merging method and number of models to use for constructing initializations (Section~\ref{subsec:mergingmethods}), and how the resulting initializations affect learning rate sensitivity (Section~\ref{subsec:lr_sensitivity}).

\subsection{Checkpoint Selection Method}
\label{subsec:checkpoint_selection}

Given a new target task, we want to quickly identify relevant checkpoints from a library without exhaustive evaluation. 
A natural candidate for such a relevance metric is the zero-shot quality of each checkpoint on the target task, which is fast to compute and requires no additional training. 
We refer to this selection strategy as \textit{Mashup checkpoint selection}, which ranks checkpoints by their zero-shot training loss or training accuracy (if available) on the target task. We use the training set rather than the validation set to avoid information leakage during selection.

To verify that this metric reliably identifies checkpoints that serve as good initializations, we compare it against two baselines: random selection and an oracle that exhaustively evaluates merges of all possible checkpoint combinations, establishing an upper bound on performance. Table~\ref{tab:checkpoint_selection_table} presents the results for Mistral-7B-Instruct-v0.2.
We observe that initializations created from checkpoints selected by Mashup consistently outperform random selection and perform comparably to the oracle. 
Furthermore, while selection by training accuracy yields slightly higher final quality, the difference compared to selection by training loss is minimal on five out of eight datasets. We attribute this to an exact match of the selected checkpoints between the two criteria on those datasets, which guides our decision to use loss-based selection in the general method as the more versatile option, since it does not require discrete labels.

\begin{table*}[t]
    \centering
    \caption{\textbf{Relevance Methods Comparison.} Accuracy results across multiple benchmarks comparing different relevance selection methods. The best result is highlighted in bold and the second best is underscored.} 
    \label{tab:checkpoint_selection_table}
    \resizebox{\linewidth}{!}{
    \begin{tabular}{llccccccccc}
        \toprule
        \textbf{Method} & \textbf{Checkpoint Selection} & \textbf{ARC-e} & \textbf{CSQA} & \textbf{Hella.} & \textbf{MathQA} & \textbf{OBQA} & \textbf{PIQA} & \textbf{SIQA} & \textbf{Wino.} & \textbf{Avg.} \\
        \midrule
        LoRA & -- & 87.32 & 83.06 & 93.38 & 21.27 & 84.38 & 86.90 & \textbf{82.79} & \underline{84.62} & 77.96 \\
        \midrule
        \multirow{4}{*}{Mashup} 
        & Random & \underline{88.42} & 82.73 & \underline{93.63} & 21.16 & 85.42 & \underline{87.77} & 32.94 & 84.54 & 72.08 \\
        & By accuracy & 88.24 & \underline{83.47} & 93.44 & \underline{40.38} & \underline{85.62} & 87.61 & 80.99 & 84.46 & \underline{80.53} \\
        & By loss & 88.24 & \underline{83.47} & 93.44 & 21.54 & \underline{85.62} & 86.95 & 81.71 & 84.46 & 78.18 \\
        & Oracle & \textbf{89.52} & \textbf{84.79} & \textbf{94.02} & \textbf{42.18} & \textbf{87.50} & \textbf{88.65} & \underline{82.63} & \textbf{84.94} & \textbf{81.78} \\
        \bottomrule
    \end{tabular}
    }
\end{table*}

\subsection{Size of dataset for checkpoint selection}
\label{subsec:eval_size}

\begin{figure}[h]
    \centering
    \includegraphics[width=\columnwidth]{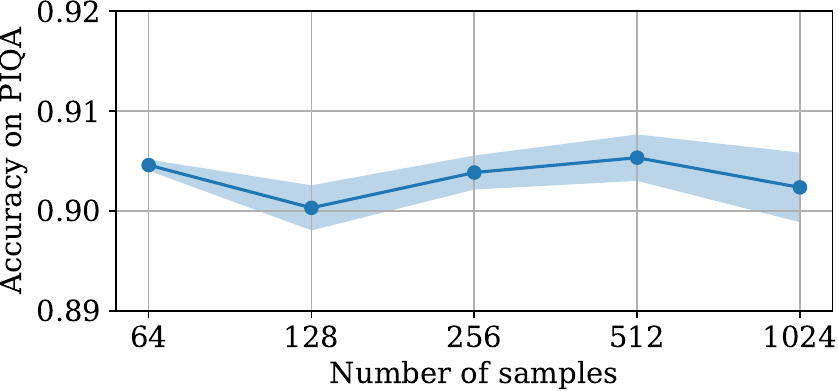}
    \caption{Accuracy on PIQA as a function of the number of samples used for checkpoint selection, evaluated on Mistral-7B-Instruct-v0.2.}
    \label{fig:eval_samples}
\end{figure}

Since the checkpoint selection step (Section~\ref{subsec:checkpoint_selection}) must be repeated for each new target task and each checkpoint in the library, we want it to be as fast as possible and investigate the minimum number of samples required for reliable selection.
We vary the number of evaluation samples, obtain the corresponding checkpoint rankings, merge the top-ranked models, and complete training from each resulting initialization.
Results are consistent across datasets, so we report only PIQA on Mistral-7B-Instruct-v0.2 in Figure~\ref{fig:eval_samples}.
While variance across runs remains relatively high, making it difficult to identify a single optimal sample size, we find that 256 samples offer a reliable trade-off: performance does not consistently improve beyond this point, and this size fits within a single evaluation batch, keeping selection efficient.

\subsection{Model Merging Method and Number of Merged Models}
\label{subsec:mergingmethods}

Initialization from a combination of several checkpoints may be beneficial, because the target task may require capabilities that were already developed while training on other tasks.
Therefore, we explore whether increasing number of checkpoints can improve performance.

Along with the naive approach of averaging the checkpoints, we evaluate the performance of several model merging techniques.
These techniques were introduced for combining multiple task-specific models into a single multi-tasking model and are effective at resolving conflicting parameters.

We hypothesize that model merging methods optimized for multi-tasking may not be optimal for model initialization before finetuning.
Therefore, we include both modern and classical methods into comparison:

\begin{itemize}[itemsep=2pt, topsep=2pt, parsep=2pt]
\item \textbf{DARE} \cite{dare}: Computes delta-parameters as the difference between each model and its initialization; randomly drops them with probability $p$; rescales the remainder by $\frac{1}{1-p}$.
\item \textbf{TIES} \cite{ties}: Computes delta-parameters as in DARE; drops smaller delta-parameters; averages delta-parameters with signs matching the majority.
\item \textbf{Fisher merging} \cite{fisher}: Computes a weighted average of parameters based on each parameter's Fisher information.
\item \textbf{RegMean} \cite{regmean}: Minimizes the difference between the predictions of the merged model and the individual models.
\end{itemize}

\begin{figure}[h]
    \centering
    \includegraphics[width=\columnwidth]{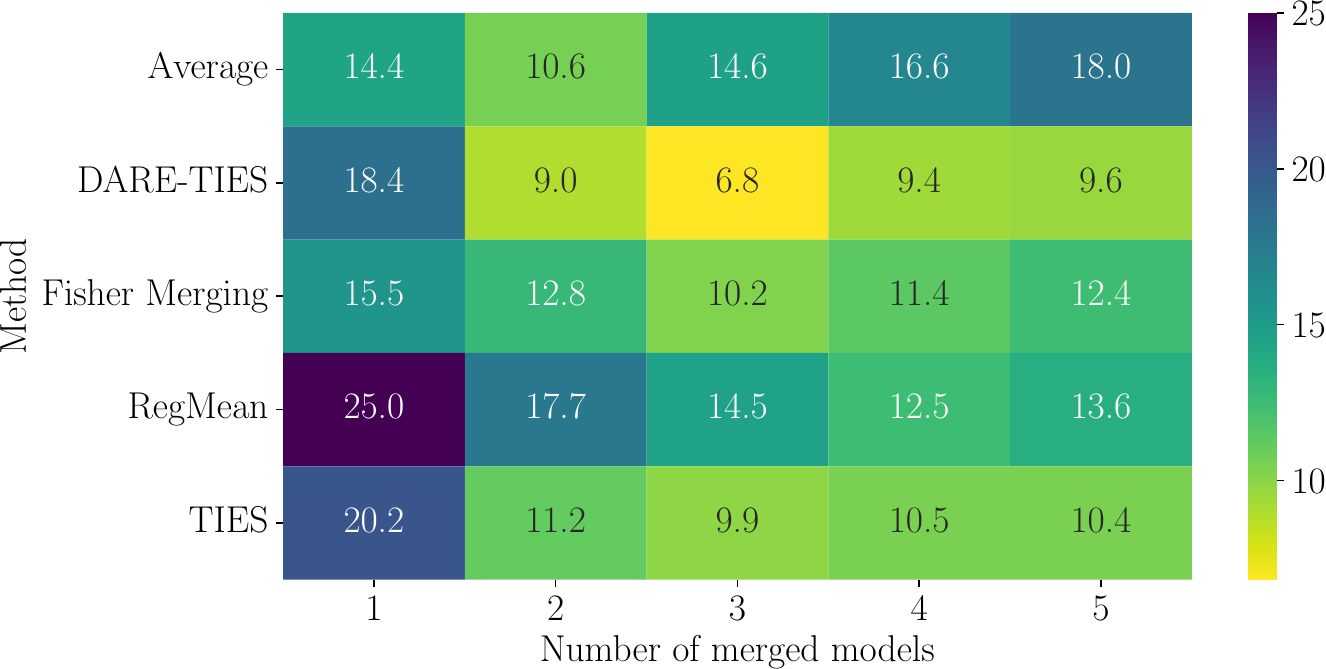}
    \caption{Mean rank by accuracy of each combination of merging method and number of merged models across 8 benchmarks (ARC-Easy, CommonsenseQA, HellaSwag, MathQA, OpenBookQA, PIQA, SocialIQA, Winogrande) in a leave-one-out setup for Mistral-7B-Instruct-v0.2.}
    \label{fig:n_merged_models}
\end{figure}

The complete results of our comparison are presented in Figure~\ref{fig:n_merged_models} for Mistral-7B-Instruct-v0.2.
Across all numbers of merged models, the combination of DARE~\cite{dare} and TIES~\cite{ties} consistently outperforms other methods.
The best overall performance is achieved by DARE-TIES when merging three models.
However, DARE-TIES requires access to the initial versions of adapters, which is not always the case when aggregating LoRA checkpoints obtained from external sources (for example, public parameter hubs).
Therefore, in practice we opt for a simpler yet effective approach: averaging the weights of the 2 most relevant models.
This simple baseline still outperforms Fisher merging and RegMean, which are impractical for our initialization use case due to both their underperformance and their additional computational and storage requirements.
Merging additional models beyond 3 provides no further benefit in our finetuning experiments, although it may be beneficial in zero-shot or low-data regimes.

Note that when only a single checkpoint is used (i.e., no merging), this reduces to sequential finetuning. As shown in Figure~\ref{fig:n_merged_models}, merging two or more checkpoints consistently outperforms this baseline.

\subsection{Learning Rate Sensitivity}
\label{subsec:lr_sensitivity}

\begin{figure}[h]
    \centering
    \includegraphics[width=\columnwidth]{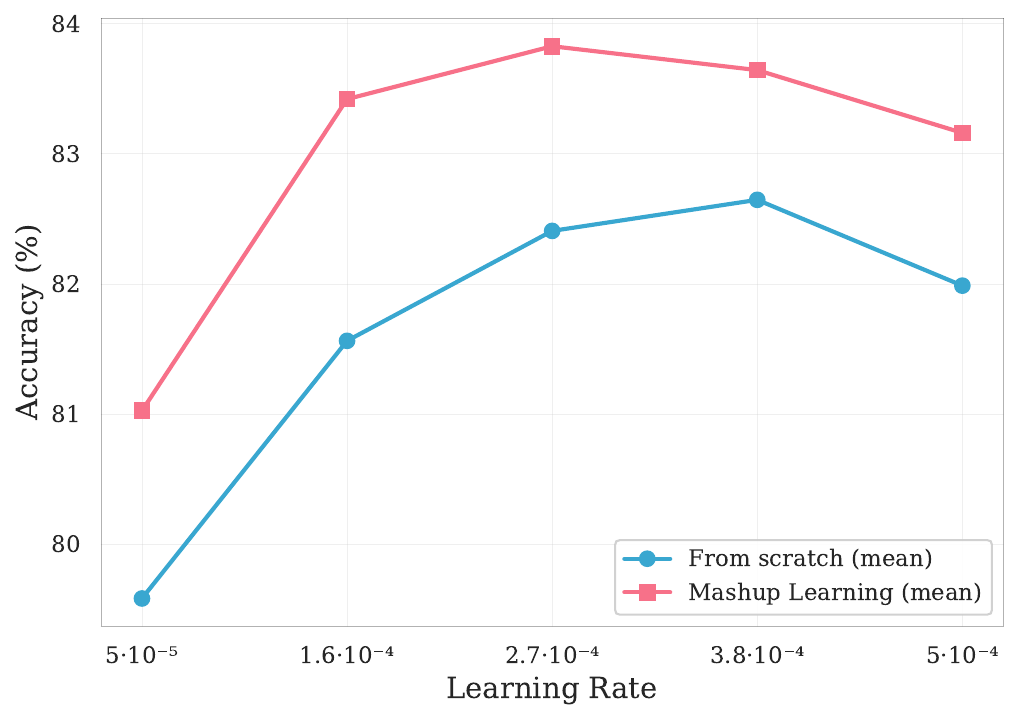}
    \caption{Gemma-3 4B LoRA sensitivity of training results to learning rates. Mean accuracy on 8 benchmarks across 3 seeds}
    \label{fig:lr_sensitivity}
\end{figure}

Selecting an appropriate learning rate typically requires extensive hyperparameter search.
The main results report accuracy at the best learning rate for each method and task.
Here, we investigate how performance varies across learning rates by sweeping five learning rates $[5\mathrm{e}{-5}, 5\mathrm{e}{-4}]$ on Gemma-3-4B and comparing training from Mashup initialization against training from scratch.
Results for LoRA are presented in Figure~\ref{fig:lr_sensitivity}.
We find that Mashup initialization consistently outperforms training from scratch across all learning rates, maintaining approximately 1.5--2 percentage points higher mean accuracy.
This suggests that Mashup initialization provides a reliable improvement regardless of learning rate choice, reducing the need for extensive hyperparameter tuning. Per-task breakdowns for this and the remaining models are provided in Appendix~\ref{app:lr_sensitivity_tasks}.

\section{Related Work}

\textbf{Model Merging}. Works related to model merging usually focus on resolving conflicts between weights of different task-specific models in order to achieve a single model capable of multi-tasking. 
Task Arithmetic \cite{taskarithmetic} introduces the concept of task vector, which is the difference between weights of a model trained on a task and a pre-trained model. 
They show it is possible to perform basic operations on such vectors and achieve multi-tasking by adding up task vectors of task-specific models. 
TIES-Merging \cite{ties} builds upon this idea. They operate on parameters of task vectors—delta parameters—discard small delta parameters, and merge only parameters shared among the majority. 
DARE \cite{dare} extends this idea too, but simplifies it significantly by dropping around 90\% of delta parameters. 
There are several distinct works related to model merging; one of them is Fisher Merging \cite{fisher}, which estimates Fisher information of each model's parameters. 
Another one, which is often used as a baseline model merging method, is RegMean \cite{regmean}. Our method does not compete in this category, as we focus on initialization rather than multi-tasking. While model merging techniques aim to combine multiple task-specific models into a single model capable of performing all tasks simultaneously, our approach uses merged checkpoints solely to initialize training for a single target task. The merged representation serves as a starting point that incorporates relevant capabilities from historical checkpoints, which is then fine-tuned specifically for the target task.
\looseness=-1

\textbf{Model Souping}. A range of works similar to model merging have emerged under the umbrella of souping. 
The key distinction is that while merging focuses on resolving conflicts between model weights, souping explores optimal proportions for combining ingredient models. 
For instance, \citet{souptogo} apply souping to mitigate catastrophic forgetting in continual learning, while \citet{modelsoup} average models trained with different hyperparameters. 
More recently, \citet{soupermodel} propose splitting tasks into weakly-correlated categories, defining experts within each category, and merging these models with specialized reweighting. 
In all these cases, the resulting soup is treated as the final model rather than as an initialization for further training.
In our paper, we explore whether combinations of historical checkpoints can serve as a better init for fine-tuning on unseen tasks.

\textbf{Zero-shot LLM adaptation}. \citet{text2lora} trained a hypernetwork that generates LoRA adapters from text descriptions of tasks, allowing the resulting adapters to be used immediately without further training. In contrast, we design our method to benefit from available training data. Our results show that Text-to-LoRA generates worse adapters than those obtained through training.

\citet{arrow} take a different approach: they select the best-suited LoRA adapter for each hidden state at every token and layer by matching hidden states to LoRA representations computed as the direction of maximum variance. However, this method imposes computational overhead during inference and is not applicable to our target use case.

\textbf{Recycling LoRAs}. Recent concurrent work by \citet{liu2026appealrealityrecyclingloras} explores a closely related setup, investigating whether practitioners can benefit from applying model merging to historical LoRA checkpoints found in the wild. 
However, their focus is on zero-shot merging as a final solution, concluding that training a vanilla LoRA from scratch outperforms merged models. 
In contrast, we use merged checkpoints as an initialization for further training, showing that this approach consistently outperforms both vanilla LoRA and zero-shot merging.

\section{Conclusion}

We presented Mashup Learning --- a method to enhance and accelerate LLM fine-tuning by selecting the most relevant historical checkpoints and aggregating them to obtain a stronger initialization.

Across three models and both LoRA and full finetuning, Mashup Learning consistently improves average accuracy by 0.5–1.9 percentage points while matching from-scratch accuracy in 41–46\% fewer training steps and up to 37\% less wall-clock time, including all overhead. The method is conceptually simple and can be viewed as a general framework for initialization through checkpoint recycling that could be extended with existing techniques.

The main contribution of our work is exploring a setup that, to our knowledge, has not been tried before and showing that it works. While the performance improvements are modest, they are consistent across different models and datasets, and the method can be easily improved further. For example, we show that applying model merging techniques to checkpoint composition improves quality, and practitioners can further experiment with model souping strategies. Task-specific refinements are also possible: we show that using accuracy instead of loss for checkpoint selection yields better results.

One limitation is that the compute savings from recycled checkpoints may be offset by the cost of estimating relevance across all source checkpoints.
Although we show that a few training batches suffice for relevance estimation, in data-constrained setups spending more compute on selection may be preferable if it improves training outcomes.

\bibliography{references}  %

@misc{kimik2,
      title={Kimi K2: Open Agentic Intelligence}, 
      author={{Kimi Team} and Yifan Bai and Yiping Bao and Y. Charles and Cheng Chen and Guanduo Chen and Haiting Chen and Huarong Chen and Jiahao Chen and Ningxin Chen and Ruijue Chen and Yanru Chen and Yuankun Chen and Yutian Chen and Zhuofu Chen and Jialei Cui and Hao Ding and Mengnan Dong and Angang Du and Chenzhuang Du and Dikang Du and Yulun Du and Yu Fan and Yichen Feng and Kelin Fu and Bofei Gao and Chenxiao Gao and Hongcheng Gao and Peizhong Gao and Tong Gao and Yuyao Ge and Shangyi Geng and Qizheng Gu and Xinran Gu and Longyu Guan and Haiqing Guo and Jianhang Guo and Xiaoru Hao and Tianhong He and Weiran He and Wenyang He and Yunjia He and Chao Hong and Hao Hu and Yangyang Hu and Zhenxing Hu and Weixiao Huang and Zhiqi Huang and Zihao Huang and Tao Jiang and Zhejun Jiang and Xinyi Jin and Yongsheng Kang and Guokun Lai and Cheng Li and Fang Li and Haoyang Li and Ming Li and Wentao Li and Yang Li and Yanhao Li and Yiwei Li and Zhaowei Li and Zheming Li and Hongzhan Lin and Xiaohan Lin and Zongyu Lin and Chengyin Liu and Chenyu Liu and Hongzhang Liu and Jingyuan Liu and Junqi Liu and Liang Liu and Shaowei Liu and T. Y. Liu and Tianwei Liu and Weizhou Liu and Yangyang Liu and Yibo Liu and Yiping Liu and Yue Liu and Zhengying Liu and Enzhe Lu and Haoyu Lu and Lijun Lu and Yashuo Luo and Shengling Ma and Xinyu Ma and Yingwei Ma and Shaoguang Mao and Jie Mei and Xin Men and Yibo Miao and Siyuan Pan and Yebo Peng and Ruoyu Qin and Zeyu Qin and Bowen Qu and Zeyu Shang and Lidong Shi and Shengyuan Shi and Feifan Song and Jianlin Su and Zhengyuan Su and Lin Sui and Xinjie Sun and Flood Sung and Yunpeng Tai and Heyi Tang and Jiawen Tao and Qifeng Teng and Chaoran Tian and Chensi Wang and Dinglu Wang and Feng Wang and Hailong Wang and Haiming Wang and Jianzhou Wang and Jiaxing Wang and Jinhong Wang and Shengjie Wang and Shuyi Wang and Si Wang and Xinyuan Wang and Yao Wang and Yejie Wang and Yiqin Wang and Yuxin Wang and Yuzhi Wang and Zhaoji Wang and Zhengtao Wang and Zhengtao Wang and Zhexu Wang and Chu Wei and Qianqian Wei and Haoning Wu and Wenhao Wu and Xingzhe Wu and Yuxin Wu and Chenjun Xiao and Jin Xie and Xiaotong Xie and Weimin Xiong and Boyu Xu and Jinjing Xu and L. H. Xu and Lin Xu and Suting Xu and Weixin Xu and Xinran Xu and Yangchuan Xu and Ziyao Xu and Jing Xu and Jing Xu and Junjie Yan and Yuzi Yan and Hao Yang and Xiaofei Yang and Yi Yang and Ying Yang and Zhen Yang and Zhilin Yang and Zonghan Yang and Haotian Yao and Xingcheng Yao and Wenjie Ye and Zhuorui Ye and Bohong Yin and Longhui Yu and Enming Yuan and Hongbang Yuan and Mengjie Yuan and Siyu Yuan and Haobing Zhan and Dehao Zhang and Hao Zhang and Wanlu Zhang and Xiaobin Zhang and Yadong Zhang and Yangkun Zhang and Yichi Zhang and Yizhi Zhang and Yongting Zhang and Yu Zhang and Yutao Zhang and Yutong Zhang and Zheng Zhang and Haotian Zhao and Yikai Zhao and Zijia Zhao and Huabin Zheng and Shaojie Zheng and Longguang Zhong and Jianren Zhou and Xinyu Zhou and Zaida Zhou and Jinguo Zhu and Zhen Zhu and Weiyu Zhuang and Xinxing Zu},
      year={2026},
      eprint={2507.20534},
      archivePrefix={arXiv},
      primaryClass={cs.LG},
      url={https://arxiv.org/abs/2507.20534}, 
}

@misc{gemma2,
      title={Gemma 2: Improving Open Language Models at a Practical Size}, 
      author={{Gemma Team} and Morgane Riviere and Shreya Pathak and Pier Giuseppe Sessa and Cassidy Hardin and Surya Bhupatiraju and Léonard Hussenot and Thomas Mesnard and Bobak Shahriari and Alexandre Ramé and Johan Ferret and Peter Liu and Pouya Tafti and Abe Friesen and Michelle Casbon and Sabela Ramos and Ravin Kumar and Charline Le Lan and Sammy Jerome and Anton Tsitsulin and Nino Vieillard and Piotr Stanczyk and Sertan Girgin and Nikola Momchev and Matt Hoffman and Shantanu Thakoor and Jean-Bastien Grill and Behnam Neyshabur and Olivier Bachem and Alanna Walton and Aliaksei Severyn and Alicia Parrish and Aliya Ahmad and Allen Hutchison and Alvin Abdagic and Amanda Carl and Amy Shen and Andy Brock and Andy Coenen and Anthony Laforge and Antonia Paterson and Ben Bastian and Bilal Piot and Bo Wu and Brandon Royal and Charlie Chen and Chintu Kumar and Chris Perry and Chris Welty and Christopher A. Choquette-Choo and Danila Sinopalnikov and David Weinberger and Dimple Vijaykumar and Dominika Rogozińska and Dustin Herbison and Elisa Bandy and Emma Wang and Eric Noland and Erica Moreira and Evan Senter and Evgenii Eltyshev and Francesco Visin and Gabriel Rasskin and Gary Wei and Glenn Cameron and Gus Martins and Hadi Hashemi and Hanna Klimczak-Plucińska and Harleen Batra and Harsh Dhand and Ivan Nardini and Jacinda Mein and Jack Zhou and James Svensson and Jeff Stanway and Jetha Chan and Jin Peng Zhou and Joana Carrasqueira and Joana Iljazi and Jocelyn Becker and Joe Fernandez and Joost van Amersfoort and Josh Gordon and Josh Lipschultz and Josh Newlan and Ju-yeong Ji and Kareem Mohamed and Kartikeya Badola and Kat Black and Katie Millican and Keelin McDonell and Kelvin Nguyen and Kiranbir Sodhia and Kish Greene and Lars Lowe Sjoesund and Lauren Usui and Laurent Sifre and Lena Heuermann and Leticia Lago and Lilly McNealus and Livio Baldini Soares and Logan Kilpatrick and Lucas Dixon and Luciano Martins and Machel Reid and Manvinder Singh and Mark Iverson and Martin Görner and Mat Velloso and Mateo Wirth and Matt Davidow and Matt Miller and Matthew Rahtz and Matthew Watson and Meg Risdal and Mehran Kazemi and Michael Moynihan and Ming Zhang and Minsuk Kahng and Minwoo Park and Mofi Rahman and Mohit Khatwani and Natalie Dao and Nenshad Bardoliwalla and Nesh Devanathan and Neta Dumai and Nilay Chauhan and Oscar Wahltinez and Pankil Botarda and Parker Barnes and Paul Barham and Paul Michel and Pengchong Jin and Petko Georgiev and Phil Culliton and Pradeep Kuppala and Ramona Comanescu and Ramona Merhej and Reena Jana and Reza Ardeshir Rokni and Rishabh Agarwal and Ryan Mullins and Samaneh Saadat and Sara Mc Carthy and Sarah Cogan and Sarah Perrin and Sébastien M. R. Arnold and Sebastian Krause and Shengyang Dai and Shruti Garg and Shruti Sheth and Sue Ronstrom and Susan Chan and Timothy Jordan and Ting Yu and Tom Eccles and Tom Hennigan and Tomas Kocisky and Tulsee Doshi and Vihan Jain and Vikas Yadav and Vilobh Meshram and Vishal Dharmadhikari and Warren Barkley and Wei Wei and Wenming Ye and Woohyun Han and Woosuk Kwon and Xiang Xu and Zhe Shen and Zhitao Gong and Zichuan Wei and Victor Cotruta and Phoebe Kirk and Anand Rao and Minh Giang and Ludovic Peran and Tris Warkentin and Eli Collins and Joelle Barral and Zoubin Ghahramani and Raia Hadsell and D. Sculley and Jeanine Banks and Anca Dragan and Slav Petrov and Oriol Vinyals and Jeff Dean and Demis Hassabis and Koray Kavukcuoglu and Clement Farabet and Elena Buchatskaya and Sebastian Borgeaud and Noah Fiedel and Armand Joulin and Kathleen Kenealy and Robert Dadashi and Alek Andreev},
      year={2024},
      eprint={2408.00118},
      archivePrefix={arXiv},
      primaryClass={cs.CL},
      url={https://arxiv.org/abs/2408.00118}, 
}

@misc{gemma3,
      title={Gemma 3 Technical Report}, 
      author={{Gemma Team} and Aishwarya Kamath and Johan Ferret and Shreya Pathak and Nino Vieillard and Ramona Merhej and Sarah Perrin and Tatiana Matejovicova and Alexandre Ramé and Morgane Rivière and Louis Rouillard and Thomas Mesnard and Geoffrey Cideron and Jean-bastien Grill and Sabela Ramos and Edouard Yvinec and Michelle Casbon and Etienne Pot and Ivo Penchev and Gaël Liu and Francesco Visin and Kathleen Kenealy and Lucas Beyer and Xiaohai Zhai and Anton Tsitsulin and Robert Busa-Fekete and Alex Feng and Noveen Sachdeva and Benjamin Coleman and Yi Gao and Basil Mustafa and Iain Barr and Emilio Parisotto and David Tian and Matan Eyal and Colin Cherry and Jan-Thorsten Peter and Danila Sinopalnikov and Surya Bhupatiraju and Rishabh Agarwal and Mehran Kazemi and Dan Malkin and Ravin Kumar and David Vilar and Idan Brusilovsky and Jiaming Luo and Andreas Steiner and Abe Friesen and Abhanshu Sharma and Abheesht Sharma and Adi Mayrav Gilady and Adrian Goedeckemeyer and Alaa Saade and Alex Feng and Alexander Kolesnikov and Alexei Bendebury and Alvin Abdagic and Amit Vadi and András György and André Susano Pinto and Anil Das and Ankur Bapna and Antoine Miech and Antoine Yang and Antonia Paterson and Ashish Shenoy and Ayan Chakrabarti and Bilal Piot and Bo Wu and Bobak Shahriari and Bryce Petrini and Charlie Chen and Charline Le Lan and Christopher A. Choquette-Choo and CJ Carey and Cormac Brick and Daniel Deutsch and Danielle Eisenbud and Dee Cattle and Derek Cheng and Dimitris Paparas and Divyashree Shivakumar Sreepathihalli and Doug Reid and Dustin Tran and Dustin Zelle and Eric Noland and Erwin Huizenga and Eugene Kharitonov and Frederick Liu and Gagik Amirkhanyan and Glenn Cameron and Hadi Hashemi and Hanna Klimczak-Plucińska and Harman Singh and Harsh Mehta and Harshal Tushar Lehri and Hussein Hazimeh and Ian Ballantyne and Idan Szpektor and Ivan Nardini and Jean Pouget-Abadie and Jetha Chan and Joe Stanton and John Wieting and Jonathan Lai and Jordi Orbay and Joseph Fernandez and Josh Newlan and Ju-yeong Ji and Jyotinder Singh and Kat Black and Kathy Yu and Kevin Hui and Kiran Vodrahalli and Klaus Greff and Linhai Qiu and Marcella Valentine and Marina Coelho and Marvin Ritter and Matt Hoffman and Matthew Watson and Mayank Chaturvedi and Michael Moynihan and Min Ma and Nabila Babar and Natasha Noy and Nathan Byrd and Nick Roy and Nikola Momchev and Nilay Chauhan and Noveen Sachdeva and Oskar Bunyan and Pankil Botarda and Paul Caron and Paul Kishan Rubenstein and Phil Culliton and Philipp Schmid and Pier Giuseppe Sessa and Pingmei Xu and Piotr Stanczyk and Pouya Tafti and Rakesh Shivanna and Renjie Wu and Renke Pan and Reza Rokni and Rob Willoughby and Rohith Vallu and Ryan Mullins and Sammy Jerome and Sara Smoot and Sertan Girgin and Shariq Iqbal and Shashir Reddy and Shruti Sheth and Siim Põder and Sijal Bhatnagar and Sindhu Raghuram Panyam and Sivan Eiger and Susan Zhang and Tianqi Liu and Trevor Yacovone and Tyler Liechty and Uday Kalra and Utku Evci and Vedant Misra and Vincent Roseberry and Vlad Feinberg and Vlad Kolesnikov and Woohyun Han and Woosuk Kwon and Xi Chen and Yinlam Chow and Yuvein Zhu and Zichuan Wei and Zoltan Egyed and Victor Cotruta and Minh Giang and Phoebe Kirk and Anand Rao and Kat Black and Nabila Babar and Jessica Lo and Erica Moreira and Luiz Gustavo Martins and Omar Sanseviero and Lucas Gonzalez and Zach Gleicher and Tris Warkentin and Vahab Mirrokni and Evan Senter and Eli Collins and Joelle Barral and Zoubin Ghahramani and Raia Hadsell and Yossi Matias and D. Sculley and Slav Petrov and Noah Fiedel and Noam Shazeer and Oriol Vinyals and Jeff Dean and Demis Hassabis and Koray Kavukcuoglu and Clement Farabet and Elena Buchatskaya and Jean-Baptiste Alayrac and Rohan Anil and Dmitry and Lepikhin and Sebastian Borgeaud and Olivier Bachem and Armand Joulin and Alek Andreev and Cassidy Hardin and Robert Dadashi and Léonard Hussenot},
      year={2025},
      eprint={2503.19786},
      archivePrefix={arXiv},
      primaryClass={cs.CL},
      url={https://arxiv.org/abs/2503.19786}, 
}

@misc{liu2026appealrealityrecyclingloras,
      title={The Appeal and Reality of Recycling LoRAs with Adaptive Merging}, 
      author={Haokun Liu and Gyung Hyun Je and Marco Ciccone and Zhenlin Xu and Prasanth YSS and Colin Raffel},
      year={2026},
      eprint={2602.12323},
      archivePrefix={arXiv},
      primaryClass={cs.LG},
      url={https://arxiv.org/abs/2602.12323}, 
}

@misc{transformer,
      title={Attention Is All You Need}, 
      author={Ashish Vaswani and Noam Shazeer and Niki Parmar and Jakob Uszkoreit and Llion Jones and Aidan N. Gomez and Lukasz Kaiser and Illia Polosukhin},
      year={2023},
      eprint={1706.03762},
      archivePrefix={arXiv},
      primaryClass={cs.CL},
      url={https://arxiv.org/abs/1706.03762}, 
}

@misc{chiang2024chatbot,
  title         = {Chatbot Arena: An Open Platform for Evaluating LLMs by Human Preference},
  author        = {Wei-Lin Chiang and Zhuohan Li and Zi Lin and Ying Sheng and Zhanghao Wu and Yonghao Zhuang and Lianmin Zheng and Siyuan Zhuang and Yingda Chen and Daliang Li and Eric P. Xing and Hao Zhang and Joseph E. Gonzalez and Ion Stoica},
  year          = {2024},
  eprint        = {2403.04132},
  archivePrefix = {arXiv},
  primaryClass  = {cs.AI}
}

@misc{lee2026learningratemattersvanilla,
      title={Learning Rate Matters: Vanilla LoRA May Suffice for LLM Fine-tuning}, 
      author={Yu-Ang Lee and Ching-Yun Ko and Pin-Yu Chen and Mi-Yen Yeh},
      year={2026},
      eprint={2602.04998},
      archivePrefix={arXiv},
      primaryClass={cs.LG},
      url={https://arxiv.org/abs/2602.04998}, 
}

@misc{modelsoup,
      title={Model soups: averaging weights of multiple fine-tuned models improves accuracy without increasing inference time}, 
      author={Mitchell Wortsman and Gabriel Ilharco and Samir Yitzhak Gadre and Rebecca Roelofs and Raphael Gontijo-Lopes and Ari S. Morcos and Hongseok Namkoong and Ali Farhadi and Yair Carmon and Simon Kornblith and Ludwig Schmidt},
      year={2022},
      eprint={2203.05482},
      archivePrefix={arXiv},
      primaryClass={cs.LG},
      url={https://arxiv.org/abs/2203.05482}, 
}

@misc{ether,
      title={ETHER: Efficient Finetuning of Large-Scale Models with Hyperplane Reflections}, 
      author={Massimo Bini and Karsten Roth and Zeynep Akata and Anna Khoreva},
      year={2024},
      eprint={2405.20271},
      archivePrefix={arXiv},
      primaryClass={cs.LG},
      url={https://arxiv.org/abs/2405.20271}, 
}

@misc{dataefficientsft,
      title={FisherSFT: Data-Efficient Supervised Fine-Tuning of Language Models Using Information Gain}, 
      author={Rohan Deb and Kiran Thekumparampil and Kousha Kalantari and Gaurush Hiranandani and Shoham Sabach and Branislav Kveton},
      year={2025},
      eprint={2505.14826},
      archivePrefix={arXiv},
      primaryClass={cs.LG},
      url={https://arxiv.org/abs/2505.14826}, 
}

@misc{qwen3,
      title={Qwen3 Technical Report}, 
      author={An Yang and Anfeng Li and Baosong Yang and Beichen Zhang and Binyuan Hui and Bo Zheng and Bowen Yu and Chang Gao and Chengen Huang and Chenxu Lv and Chujie Zheng and Dayiheng Liu and Fan Zhou and Fei Huang and Feng Hu and Hao Ge and Haoran Wei and Huan Lin and Jialong Tang and Jian Yang and Jianhong Tu and Jianwei Zhang and Jianxin Yang and Jiaxi Yang and Jing Zhou and Jingren Zhou and Junyang Lin and Kai Dang and Keqin Bao and Kexin Yang and Le Yu and Lianghao Deng and Mei Li and Mingfeng Xue and Mingze Li and Pei Zhang and Peng Wang and Qin Zhu and Rui Men and Ruize Gao and Shixuan Liu and Shuang Luo and Tianhao Li and Tianyi Tang and Wenbiao Yin and Xingzhang Ren and Xinyu Wang and Xinyu Zhang and Xuancheng Ren and Yang Fan and Yang Su and Yichang Zhang and Yinger Zhang and Yu Wan and Yuqiong Liu and Zekun Wang and Zeyu Cui and Zhenru Zhang and Zhipeng Zhou and Zihan Qiu},
      year={2025},
      eprint={2505.09388},
      archivePrefix={arXiv},
      primaryClass={cs.CL},
      url={https://arxiv.org/abs/2505.09388}, 
}

@misc{llama3,
      title={The Llama 3 Herd of Models}, 
      author={Aaron Grattafiori and Abhimanyu Dubey and Abhinav Jauhri and Abhinav Pandey and Abhishek Kadian and Ahmad Al-Dahle and Aiesha Letman and Akhil Mathur and Alan Schelten and Alex Vaughan and Amy Yang and Angela Fan and Anirudh Goyal and Anthony Hartshorn and Aobo Yang and Archi Mitra and Archie Sravankumar and Artem Korenev and Arthur Hinsvark and Arun Rao and Aston Zhang and Aurelien Rodriguez and Austen Gregerson and Ava Spataru and Baptiste Roziere and Bethany Biron and Binh Tang and Bobbie Chern and Charlotte Caucheteux and Chaya Nayak and Chloe Bi and Chris Marra and Chris McConnell and Christian Keller and Christophe Touret and Chunyang Wu and Corinne Wong and Cristian Canton Ferrer and Cyrus Nikolaidis and Damien Allonsius and Daniel Song and Danielle Pintz and Danny Livshits and Danny Wyatt and David Esiobu and Dhruv Choudhary and Dhruv Mahajan and Diego Garcia-Olano and Diego Perino and Dieuwke Hupkes and Egor Lakomkin and Ehab AlBadawy and Elina Lobanova and Emily Dinan and Eric Michael Smith and Filip Radenovic and Francisco Guzmán and Frank Zhang and Gabriel Synnaeve and Gabrielle Lee and Georgia Lewis Anderson and Govind Thattai and Graeme Nail and Gregoire Mialon and Guan Pang and Guillem Cucurell and Hailey Nguyen and Hannah Korevaar and Hu Xu and Hugo Touvron and Iliyan Zarov and Imanol Arrieta Ibarra and Isabel Kloumann and Ishan Misra and Ivan Evtimov and Jack Zhang and Jade Copet and Jaewon Lee and Jan Geffert and Jana Vranes and Jason Park and Jay Mahadeokar and Jeet Shah and Jelmer van der Linde and Jennifer Billock and Jenny Hong and Jenya Lee and Jeremy Fu and Jianfeng Chi and Jianyu Huang and Jiawen Liu and Jie Wang and Jiecao Yu and Joanna Bitton and Joe Spisak and Jongsoo Park and Joseph Rocca and Joshua Johnstun and Joshua Saxe and Junteng Jia and Kalyan Vasuden Alwala and Karthik Prasad and Kartikeya Upasani and Kate Plawiak and Ke Li and Kenneth Heafield and Kevin Stone and Khalid El-Arini and Krithika Iyer and Kshitiz Malik and Kuenley Chiu and Kunal Bhalla and Kushal Lakhotia and Lauren Rantala-Yeary and Laurens van der Maaten and Lawrence Chen and Liang Tan and Liz Jenkins and Louis Martin and Lovish Madaan and Lubo Malo and Lukas Blecher and Lukas Landzaat and Luke de Oliveira and Madeline Muzzi and Mahesh Pasupuleti and Mannat Singh and Manohar Paluri and Marcin Kardas and Maria Tsimpoukelli and Mathew Oldham and Mathieu Rita and Maya Pavlova and Melanie Kambadur and Mike Lewis and Min Si and Mitesh Kumar Singh and Mona Hassan and Naman Goyal and Narjes Torabi and Nikolay Bashlykov and Nikolay Bogoychev and Niladri Chatterji and Ning Zhang and Olivier Duchenne and Onur Çelebi and Patrick Alrassy and Pengchuan Zhang and Pengwei Li and Petar Vasic and Peter Weng and Prajjwal Bhargava and Pratik Dubal and Praveen Krishnan and Punit Singh Koura and Puxin Xu and Qing He and Qingxiao Dong and Ragavan Srinivasan and Raj Ganapathy and Ramon Calderer and Ricardo Silveira Cabral and Robert Stojnic and Roberta Raileanu and Rohan Maheswari and Rohit Girdhar and Rohit Patel and Romain Sauvestre and Ronnie Polidoro and Roshan Sumbaly and Ross Taylor and Ruan Silva and Rui Hou and Rui Wang and Saghar Hosseini and Sahana Chennabasappa and Sanjay Singh and Sean Bell and Seohyun Sonia Kim and Sergey Edunov and Shaoliang Nie and Sharan Narang and Sharath Raparthy and Sheng Shen and Shengye Wan and Shruti Bhosale and Shun Zhang and Simon Vandenhende and Soumya Batra and Spencer Whitman and Sten Sootla and Stephane Collot and Suchin Gururangan and Sydney Borodinsky and Tamar Herman and Tara Fowler and Tarek Sheasha and Thomas Georgiou and Thomas Scialom and Tobias Speckbacher and Todor Mihaylov and Tong Xiao and Ujjwal Karn and Vedanuj Goswami and Vibhor Gupta and Vignesh Ramanathan and Viktor Kerkez and Vincent Gonguet and Virginie Do and Vish Vogeti and Vítor Albiero and Vladan Petrovic and Weiwei Chu and Wenhan Xiong and Wenyin Fu and Whitney Meers and Xavier Martinet and Xiaodong Wang and Xiaofang Wang and Xiaoqing Ellen Tan and Xide Xia and Xinfeng Xie and Xuchao Jia and Xuewei Wang and Yaelle Goldschlag and Yashesh Gaur and Yasmine Babaei and Yi Wen and Yiwen Song and Yuchen Zhang and Yue Li and Yuning Mao and Zacharie Delpierre Coudert and Zheng Yan and Zhengxing Chen and Zoe Papakipos and Aaditya Singh and Aayushi Srivastava and Abha Jain and Adam Kelsey and Adam Shajnfeld and Adithya Gangidi and Adolfo Victoria and Ahuva Goldstand and Ajay Menon and Ajay Sharma and Alex Boesenberg and Alexei Baevski and Allie Feinstein and Amanda Kallet and Amit Sangani and Amos Teo and Anam Yunus and Andrei Lupu and Andres Alvarado and Andrew Caples and Andrew Gu and Andrew Ho and Andrew Poulton and Andrew Ryan and Ankit Ramchandani and Annie Dong and Annie Franco and Anuj Goyal and Aparajita Saraf and Arkabandhu Chowdhury and Ashley Gabriel and Ashwin Bharambe and Assaf Eisenman and Azadeh Yazdan and Beau James and Ben Maurer and Benjamin Leonhardi and Bernie Huang and Beth Loyd and Beto De Paola and Bhargavi Paranjape and Bing Liu and Bo Wu and Boyu Ni and Braden Hancock and Bram Wasti and Brandon Spence and Brani Stojkovic and Brian Gamido and Britt Montalvo and Carl Parker and Carly Burton and Catalina Mejia and Ce Liu and Changhan Wang and Changkyu Kim and Chao Zhou and Chester Hu and Ching-Hsiang Chu and Chris Cai and Chris Tindal and Christoph Feichtenhofer and Cynthia Gao and Damon Civin and Dana Beaty and Daniel Kreymer and Daniel Li and David Adkins and David Xu and Davide Testuggine and Delia David and Devi Parikh and Diana Liskovich and Didem Foss and Dingkang Wang and Duc Le and Dustin Holland and Edward Dowling and Eissa Jamil and Elaine Montgomery and Eleonora Presani and Emily Hahn and Emily Wood and Eric-Tuan Le and Erik Brinkman and Esteban Arcaute and Evan Dunbar and Evan Smothers and Fei Sun and Felix Kreuk and Feng Tian and Filippos Kokkinos and Firat Ozgenel and Francesco Caggioni and Frank Kanayet and Frank Seide and Gabriela Medina Florez and Gabriella Schwarz and Gada Badeer and Georgia Swee and Gil Halpern and Grant Herman and Grigory Sizov and Guangyi and Zhang and Guna Lakshminarayanan and Hakan Inan and Hamid Shojanazeri and Han Zou and Hannah Wang and Hanwen Zha and Haroun Habeeb and Harrison Rudolph and Helen Suk and Henry Aspegren and Hunter Goldman and Hongyuan Zhan and Ibrahim Damlaj and Igor Molybog and Igor Tufanov and Ilias Leontiadis and Irina-Elena Veliche and Itai Gat and Jake Weissman and James Geboski and James Kohli and Janice Lam and Japhet Asher and Jean-Baptiste Gaya and Jeff Marcus and Jeff Tang and Jennifer Chan and Jenny Zhen and Jeremy Reizenstein and Jeremy Teboul and Jessica Zhong and Jian Jin and Jingyi Yang and Joe Cummings and Jon Carvill and Jon Shepard and Jonathan McPhie and Jonathan Torres and Josh Ginsburg and Junjie Wang and Kai Wu and Kam Hou U and Karan Saxena and Kartikay Khandelwal and Katayoun Zand and Kathy Matosich and Kaushik Veeraraghavan and Kelly Michelena and Keqian Li and Kiran Jagadeesh and Kun Huang and Kunal Chawla and Kyle Huang and Lailin Chen and Lakshya Garg and Lavender A and Leandro Silva and Lee Bell and Lei Zhang and Liangpeng Guo and Licheng Yu and Liron Moshkovich and Luca Wehrstedt and Madian Khabsa and Manav Avalani and Manish Bhatt and Martynas Mankus and Matan Hasson and Matthew Lennie and Matthias Reso and Maxim Groshev and Maxim Naumov and Maya Lathi and Meghan Keneally and Miao Liu and Michael L. Seltzer and Michal Valko and Michelle Restrepo and Mihir Patel and Mik Vyatskov and Mikayel Samvelyan and Mike Clark and Mike Macey and Mike Wang and Miquel Jubert Hermoso and Mo Metanat and Mohammad Rastegari and Munish Bansal and Nandhini Santhanam and Natascha Parks and Natasha White and Navyata Bawa and Nayan Singhal and Nick Egebo and Nicolas Usunier and Nikhil Mehta and Nikolay Pavlovich Laptev and Ning Dong and Norman Cheng and Oleg Chernoguz and Olivia Hart and Omkar Salpekar and Ozlem Kalinli and Parkin Kent and Parth Parekh and Paul Saab and Pavan Balaji and Pedro Rittner and Philip Bontrager and Pierre Roux and Piotr Dollar and Polina Zvyagina and Prashant Ratanchandani and Pritish Yuvraj and Qian Liang and Rachad Alao and Rachel Rodriguez and Rafi Ayub and Raghotham Murthy and Raghu Nayani and Rahul Mitra and Rangaprabhu Parthasarathy and Raymond Li and Rebekkah Hogan and Robin Battey and Rocky Wang and Russ Howes and Ruty Rinott and Sachin Mehta and Sachin Siby and Sai Jayesh Bondu and Samyak Datta and Sara Chugh and Sara Hunt and Sargun Dhillon and Sasha Sidorov and Satadru Pan and Saurabh Mahajan and Saurabh Verma and Seiji Yamamoto and Sharadh Ramaswamy and Shaun Lindsay and Shaun Lindsay and Sheng Feng and Shenghao Lin and Shengxin Cindy Zha and Shishir Patil and Shiva Shankar and Shuqiang Zhang and Shuqiang Zhang and Sinong Wang and Sneha Agarwal and Soji Sajuyigbe and Soumith Chintala and Stephanie Max and Stephen Chen and Steve Kehoe and Steve Satterfield and Sudarshan Govindaprasad and Sumit Gupta and Summer Deng and Sungmin Cho and Sunny Virk and Suraj Subramanian and Sy Choudhury and Sydney Goldman and Tal Remez and Tamar Glaser and Tamara Best and Thilo Koehler and Thomas Robinson and Tianhe Li and Tianjun Zhang and Tim Matthews and Timothy Chou and Tzook Shaked and Varun Vontimitta and Victoria Ajayi and Victoria Montanez and Vijai Mohan and Vinay Satish Kumar and Vishal Mangla and Vlad Ionescu and Vlad Poenaru and Vlad Tiberiu Mihailescu and Vladimir Ivanov and Wei Li and Wenchen Wang and Wenwen Jiang and Wes Bouaziz and Will Constable and Xiaocheng Tang and Xiaojian Wu and Xiaolan Wang and Xilun Wu and Xinbo Gao and Yaniv Kleinman and Yanjun Chen and Ye Hu and Ye Jia and Ye Qi and Yenda Li and Yilin Zhang and Ying Zhang and Yossi Adi and Youngjin Nam and Yu and Wang and Yu Zhao and Yuchen Hao and Yundi Qian and Yunlu Li and Yuzi He and Zach Rait and Zachary DeVito and Zef Rosnbrick and Zhaoduo Wen and Zhenyu Yang and Zhiwei Zhao and Zhiyu Ma},
      year={2024},
      eprint={2407.21783},
      archivePrefix={arXiv},
      primaryClass={cs.AI},
      url={https://arxiv.org/abs/2407.21783}, 
}

@misc{peft,
      title={Parameter-Efficient Fine-Tuning Methods for Pretrained Language Models: A Critical Review and Assessment}, 
      author={Lingling Xu and Haoran Xie and Si-Zhao Joe Qin and Xiaohui Tao and Fu Lee Wang},
      year={2023},
      eprint={2312.12148},
      archivePrefix={arXiv},
      primaryClass={cs.CL},
      url={https://arxiv.org/abs/2312.12148}, 
}

@misc{lora,
      title={LoRA: Low-Rank Adaptation of Large Language Models}, 
      author={Edward J. Hu and Yelong Shen and Phillip Wallis and Zeyuan Allen-Zhu and Yuanzhi Li and Shean Wang and Lu Wang and Weizhu Chen},
      year={2021},
      eprint={2106.09685},
      archivePrefix={arXiv},
      primaryClass={cs.CL},
      url={https://arxiv.org/abs/2106.09685}, 
}

@misc{doremi,
      title={DoReMi: Optimizing Data Mixtures Speeds Up Language Model Pretraining}, 
      author={Sang Michael Xie and Hieu Pham and Xuanyi Dong and Nan Du and Hanxiao Liu and Yifeng Lu and Percy Liang and Quoc V. Le and Tengyu Ma and Adams Wei Yu},
      year={2023},
      eprint={2305.10429},
      archivePrefix={arXiv},
      primaryClass={cs.CL},
      url={https://arxiv.org/abs/2305.10429}, 
}

@misc{domain2vec,
      title={Domain2Vec: Vectorizing Datasets to Find the Optimal Data Mixture without Training}, 
      author={Mozhi Zhang and Howe Tissue and Lu Wang and Xipeng Qiu},
      year={2025},
      eprint={2506.10952},
      archivePrefix={arXiv},
      primaryClass={cs.CL},
      url={https://arxiv.org/abs/2506.10952}, 
}

@misc{arrow,
      title={Towards Modular LLMs by Building and Reusing a Library of LoRAs}, 
      author={Oleksiy Ostapenko and Zhan Su and Edoardo Maria Ponti and Laurent Charlin and Nicolas Le Roux and Matheus Pereira and Lucas Caccia and Alessandro Sordoni},
      year={2024},
      eprint={2405.11157},
      archivePrefix={arXiv},
      primaryClass={cs.LG},
      url={https://arxiv.org/abs/2405.11157}, 
}

@misc{souptogo,
      title={Soup to go: mitigating forgetting during continual learning with model averaging}, 
      author={Anat Kleiman and Gintare Karolina Dziugaite and Jonathan Frankle and Sham Kakade and Mansheej Paul},
      year={2025},
      eprint={2501.05559},
      archivePrefix={arXiv},
      primaryClass={cs.LG},
      url={https://arxiv.org/abs/2501.05559}, 
}

@misc{taskarithmetic,
      title={Editing Models with Task Arithmetic}, 
      author={Gabriel Ilharco and Marco Tulio Ribeiro and Mitchell Wortsman and Suchin Gururangan and Ludwig Schmidt and Hannaneh Hajishirzi and Ali Farhadi},
      year={2023},
      eprint={2212.04089},
      archivePrefix={arXiv},
      primaryClass={cs.LG},
      url={https://arxiv.org/abs/2212.04089}, 
}

@misc{text2lora,
      title={Text-to-LoRA: Instant Transformer Adaption}, 
      author={Rujikorn Charakorn and Edoardo Cetin and Yujin Tang and Robert Tjarko Lange},
      year={2025},
      eprint={2506.06105},
      archivePrefix={arXiv},
      primaryClass={cs.LG},
      url={https://arxiv.org/abs/2506.06105}, 
}

@article{lorawithoutregret,
  author = {John Schulman and {Thinking Machines Lab}},
  title = {LoRA Without Regret},
  journal = {Thinking Machines Lab: Connectionism},
  year = {2025},
  note = {\url{https://thinkingmachines.ai/blog/lora/}},
  doi = {10.64434/tml.20250929},
}

@misc{regmean,
      title={Dataless Knowledge Fusion by Merging Weights of Language Models}, 
      author={Xisen Jin and Xiang Ren and Daniel Preotiuc-Pietro and Pengxiang Cheng},
      year={2025},
      eprint={2212.09849},
      archivePrefix={arXiv},
      primaryClass={cs.CL},
      url={https://arxiv.org/abs/2212.09849}, 
}

@misc{dare,
      title={Language Models are Super Mario: Absorbing Abilities from Homologous Models as a Free Lunch}, 
      author={Le Yu and Bowen Yu and Haiyang Yu and Fei Huang and Yongbin Li},
      year={2024},
      eprint={2311.03099},
      archivePrefix={arXiv},
      primaryClass={cs.CL},
      url={https://arxiv.org/abs/2311.03099}, 
}

@misc{soupermodel,
      title={Souper-Model: How Simple Arithmetic Unlocks State-of-the-Art LLM Performance}, 
      author={Shalini Maiti and Amar Budhiraja and Bhavul Gauri and Gaurav Chaurasia and Anton Protopopov and Alexis Audran-Reiss and Michael Slater and Despoina Magka and Tatiana Shavrina and Roberta Raileanu and Yoram Bachrach},
      year={2025},
      eprint={2511.13254},
      archivePrefix={arXiv},
      primaryClass={cs.CL},
      url={https://arxiv.org/abs/2511.13254}, 
}

@misc{ties,
      title={TIES-Merging: Resolving Interference When Merging Models}, 
      author={Prateek Yadav and Derek Tam and Leshem Choshen and Colin Raffel and Mohit Bansal},
      year={2023},
      eprint={2306.01708},
      archivePrefix={arXiv},
      primaryClass={cs.LG},
      url={https://arxiv.org/abs/2306.01708}, 
}

@misc{fisher,
      title={Merging Models with Fisher-Weighted Averaging}, 
      author={Michael Matena and Colin Raffel},
      year={2022},
      eprint={2111.09832},
      archivePrefix={arXiv},
      primaryClass={cs.LG},
      url={https://arxiv.org/abs/2111.09832}, 
}

@misc{loralearns,
  title={LoRA Learns Less and Forgets Less},
  author={Dan Biderman and Jacob Portes and Jose Javier Gonzalez Ortiz and Mansheej Paul and Philip Greengard and Connor Jennings and Daniel King and Sam Havens and Vitaliy Chiley and Jonathan Frankle and Cody Blakeney and John Patrick Cunningham},
  year={2024},
  eprint={2405.09673},
  archivePrefix={arXiv},
  primaryClass={cs.LG},
  url={https://arxiv.org/abs/2405.09673}
}

@inproceedings{clark2018think,
  title={Think you have Solved Question Answering? Try ARC, the AI2 Reasoning Challenge},
  author={Clark, Peter and Cowhey, Isaac and Etzioni, Oren and Khot, Tushar and Sabharwal, Ashish and Schoenick, Carissa and Tafjord, Oyvind},
  booktitle={Proceedings of the 2018 Conference on Empirical Methods in Natural Language Processing (EMNLP)},
  pages={4459--4470},
  year={2018}
}

@inproceedings{mihaylov2018openbookqa,
  title={OpenBookQA: A "Can a first-grader answer this question?" style QA dataset},
  author={Mihaylov, Todor and Clark, Peter and Khot, Tushar and Sabharwal, Ashish},
  booktitle={Proceedings of the 2018 Conference on Empirical Methods in Natural Language Processing (EMNLP)},
  pages={5377--5382},
  year={2018}
}

@inproceedings{sakaguchi2019winogrande,
  title={WinoGrande: An Adversarial Winograd Schema Challenge at Scale},
  author={Sakaguchi, Keisuke and Bras, Ronan Le and Todorovic, Mali and Bhagavatula, Chandra and Choi, Yejin},
  booktitle={Proceedings of the 2019 Conference on Empirical Methods in Natural Language Processing and the 9th International Joint Conference on Natural Language Processing (EMNLP-IJCNLP)},
  pages={5255--5266},
  year={2019}
}

@inproceedings{bisk2020piqa,
  title={PIQA: Reasoning about Physical Commonsense in Natural Language},
  author={Bisk, Yonatan and Zellers, Rowan and Goyal, Naman and Choi, Yejin and Hockenmaier, Julia},
  booktitle={Proceedings of the 34th AAAI Conference on Artificial Intelligence (AAAI)},
  volume={34},
  number={05},
  pages={9433--9440},
  year={2020}
}

@inproceedings{amini2019mathqa,
  title={MathQA: Towards Interpretable Math Word Problem Solving with Operation-Based Formalisms},
  author={Amini, Aida and Gabriel, Saadia and Lin, Shanchuan and Kohli, Peter and Susskind, Joshua and Hooker, Sara},
  booktitle={Proceedings of the 2019 Conference on Empirical Methods in Natural Language Processing and the 9th International Joint Conference on Natural Language Processing (EMNLP-IJCNLP)},
  pages={5447--5457},
  year={2019}
}

@inproceedings{zellers2019hellaswag,
  title={HellaSwag: Can a Machine Really Finish Your Sentence?},
  author={Zellers, Rowan and Holtzman, Ari and Bisk, Yonatan and Farhadi, Ali and Choi, Yejin},
  booktitle={Proceedings of the 57th Annual Meeting of the Association for Computational Linguistics (ACL)},
  pages={4791--4800},
  year={2019}
}

@inproceedings{sap2019socialiqa,
  title={SocialIQa: Commonsense Reasoning about Social Interactions},
  author={Sap, Maarten and Le Bras, Ronan and Allaway, Emily and Bhagavatula, Chandra and Lourie, Nicholas and Rashkin, Hannah and Roof, Brendan and Smith, Noah A and Choi, Yejin},
  booktitle={Proceedings of the 2019 Conference on Empirical Methods in Natural Language Processing and the 9th International Joint Conference on Natural Language Processing (EMNLP-IJCNLP)},
  pages={1253--1265},
  year={2019}
}

@inproceedings{talmor2019commonsenseqa,
  title={CommonsenseQA: A Question Answering Challenge Targeting Commonsense Knowledge},
  author={Talmor, Alon and Herzig, Jonathan and Lourie, Nicholas and Berant, Jonathan},
  booktitle={Proceedings of the 2019 Conference of the North American Chapter of the Association for Computational Linguistics: Human Language Technologies (NAACL-HLT)},
  volume={1},
  pages={4149--4158},
  year={2019}
}

@misc{jiang2023mistral7b,
      title={Mistral 7B}, 
      author={Albert Q. Jiang and Alexandre Sablayrolles and Arthur Mensch and Chris Bamford and Devendra Singh Chaplot and Diego de las Casas and Florian Bressand and Gianna Lengyel and Guillaume Lample and Lucile Saulnier and Lélio Renard Lavaud and Marie-Anne Lachaux and Pierre Stock and Teven Le Scao and Thibaut Lavril and Thomas Wang and Timothée Lacroix and William El Sayed},
      year={2023},
      eprint={2310.06825},
      archivePrefix={arXiv},
      primaryClass={cs.CL},
      url={https://arxiv.org/abs/2310.06825}, 
}

@misc{brüelgabrielsson2025compressserveservingthousands,
      title={Compress then Serve: Serving Thousands of LoRA Adapters with Little Overhead}, 
      author={Rickard Brüel-Gabrielsson and Jiacheng Zhu and Onkar Bhardwaj and Leshem Choshen and Kristjan Greenewald and Mikhail Yurochkin and Justin Solomon},
      year={2025},
      eprint={2407.00066},
      archivePrefix={arXiv},
      primaryClass={cs.DC},
      url={https://arxiv.org/abs/2407.00066}, 
}

@misc{kaushik2025universalweightsubspacehypothesis,
      title={The Universal Weight Subspace Hypothesis}, 
      author={Prakhar Kaushik and Shravan Chaudhari and Ankit Vaidya and Rama Chellappa and Alan Yuille},
      year={2025},
      eprint={2512.05117},
      archivePrefix={arXiv},
      primaryClass={cs.LG},
      url={https://arxiv.org/abs/2512.05117}, 
}

@inproceedings{howard-ruder-2018-universal,
    title = "Universal Language Model Fine-tuning for Text Classification",
    author = "Howard, Jeremy  and
      Ruder, Sebastian",
    editor = "Gurevych, Iryna  and
      Miyao, Yusuke",
    booktitle = "Proceedings of the 56th Annual Meeting of the Association for Computational Linguistics (Volume 1: Long Papers)",
    month = jul,
    year = "2018",
    address = "Melbourne, Australia",
    publisher = "Association for Computational Linguistics",
    url = "https://aclanthology.org/P18-1031/",
    doi = "10.18653/v1/P18-1031",
    pages = "328--339"
}

@misc{olmo2025olmo3,
      title={Olmo 3}, 
      author={{Team Olmo} and Allyson Ettinger and Amanda Bertsch and Bailey Kuehl and David Graham and David Heineman and Dirk Groeneveld and Faeze Brahman and Finbarr Timbers and Hamish Ivison and Jacob Morrison and Jake Poznanski and Kyle Lo and Luca Soldaini and Matt Jordan and Mayee Chen and Michael Noukhovitch and Nathan Lambert and Pete Walsh and Pradeep Dasigi and Robert Berry and Saumya Malik and Saurabh Shah and Scott Geng and Shane Arora and Shashank Gupta and Taira Anderson and Teng Xiao and Tyler Murray and Tyler Romero and Victoria Graf and Akari Asai and Akshita Bhagia and Alexander Wettig and Alisa Liu and Aman Rangapur and Chloe Anastasiades and Costa Huang and Dustin Schwenk and Harsh Trivedi and Ian Magnusson and Jaron Lochner and Jiacheng Liu and Lester James V. Miranda and Maarten Sap and Malia Morgan and Michael Schmitz and Michal Guerquin and Michael Wilson and Regan Huff and Ronan Le Bras and Rui Xin and Rulin Shao and Sam Skjonsberg and Shannon Zejiang Shen and Shuyue Stella Li and Tucker Wilde and Valentina Pyatkin and Will Merrill and Yapei Chang and Yuling Gu and Zhiyuan Zeng and Ashish Sabharwal and Luke Zettlemoyer and Pang Wei Koh and Ali Farhadi and Noah A. Smith and Hannaneh Hajishirzi},
      year={2025},
      eprint={2512.13961},
      archivePrefix={arXiv},
      primaryClass={cs.CL},
      url={https://arxiv.org/abs/2512.13961}, 
}

@misc{cohere2025commandaenterprisereadylarge,
      title={Command A: An Enterprise-Ready Large Language Model}, 
      author={Team Cohere and : and Aakanksha and Arash Ahmadian and Marwan Ahmed and Jay Alammar and Milad Alizadeh and Yazeed Alnumay and Sophia Althammer and Arkady Arkhangorodsky and Viraat Aryabumi and Dennis Aumiller and Raphaël Avalos and Zahara Aviv and Sammie Bae and Saurabh Baji and Alexandre Barbet and Max Bartolo and Björn Bebensee and Neeral Beladia and Walter Beller-Morales and Alexandre Bérard and Andrew Berneshawi and Anna Bialas and Phil Blunsom and Matt Bobkin and Adi Bongale and Sam Braun and Maxime Brunet and Samuel Cahyawijaya and David Cairuz and Jon Ander Campos and Cassie Cao and Kris Cao and Roman Castagné and Julián Cendrero and Leila Chan Currie and Yash Chandak and Diane Chang and Giannis Chatziveroglou and Hongyu Chen and Claire Cheng and Alexis Chevalier and Justin T. Chiu and Eugene Cho and Eugene Choi and Eujeong Choi and Tim Chung and Volkan Cirik and Ana Cismaru and Pierre Clavier and Henry Conklin and Lucas Crawhall-Stein and Devon Crouse and Andres Felipe Cruz-Salinas and Ben Cyrus and Daniel D'souza and Hugo Dalla-Torre and John Dang and William Darling and Omar Darwiche Domingues and Saurabh Dash and Antoine Debugne and Théo Dehaze and Shaan Desai and Joan Devassy and Rishit Dholakia and Kyle Duffy and Ali Edalati and Ace Eldeib and Abdullah Elkady and Sarah Elsharkawy and Irem Ergün and Beyza Ermis and Marzieh Fadaee and Boyu Fan and Lucas Fayoux and Yannis Flet-Berliac and Nick Frosst and Matthias Gallé and Wojciech Galuba and Utsav Garg and Matthieu Geist and Mohammad Gheshlaghi Azar and Ellen Gilsenan-McMahon and Seraphina Goldfarb-Tarrant and Tomas Goldsack and Aidan Gomez and Victor Machado Gonzaga and Nithya Govindarajan and Manoj Govindassamy and Nathan Grinsztajn and Nikolas Gritsch and Patrick Gu and Shangmin Guo and Kilian Haefeli and Rod Hajjar and Tim Hawes and Jingyi He and Sebastian Hofstätter and Sungjin Hong and Sara Hooker and Tom Hosking and Stephanie Howe and Eric Hu and Renjie Huang and Hemant Jain and Ritika Jain and Nick Jakobi and Madeline Jenkins and JJ Jordan and Dhruti Joshi and Jason Jung and Trushant Kalyanpur and Siddhartha Rao Kamalakara and Julia Kedrzycki and Gokce Keskin and Edward Kim and Joon Kim and Wei-Yin Ko and Tom Kocmi and Michael Kozakov and Wojciech Kryściński and Arnav Kumar Jain and Komal Kumar Teru and Sander Land and Michael Lasby and Olivia Lasche and Justin Lee and Patrick Lewis and Jeffrey Li and Jonathan Li and Hangyu Lin and Acyr Locatelli and Kevin Luong and Raymond Ma and Lukáš Mach and Marina Machado and Joanne Magbitang and Brenda Malacara Lopez and Aryan Mann and Kelly Marchisio and Olivia Markham and Alexandre Matton and Alex McKinney and Dominic McLoughlin and Jozef Mokry and Adrien Morisot and Autumn Moulder and Harry Moynehan and Maximilian Mozes and Vivek Muppalla and Lidiya Murakhovska and Hemangani Nagarajan and Alekhya Nandula and Hisham Nasir and Shauna Nehra and Josh Netto-Rosen and Daniel Ohashi and James Owers-Bardsley and Jason Ozuzu and Dennis Padilla and Gloria Park and Sam Passaglia and Jeremy Pekmez and Laura Penstone and Aleksandra Piktus and Case Ploeg and Andrew Poulton and Youran Qi and Shubha Raghvendra and Miguel Ramos and Ekagra Ranjan and Pierre Richemond and Cécile Robert-Michon and Aurélien Rodriguez and Sudip Roy and Sebastian Ruder and Laura Ruis and Louise Rust and Anubhav Sachan and Alejandro Salamanca and Kailash Karthik Saravanakumar and Isha Satyakam and Alice Schoenauer Sebag and Priyanka Sen and Sholeh Sepehri and Preethi Seshadri and Ye Shen and Tom Sherborne and Sylvie Shang Shi and Sanal Shivaprasad and Vladyslav Shmyhlo and Anirudh Shrinivason and Inna Shteinbuk and Amir Shukayev and Mathieu Simard and Ella Snyder and Ava Spataru and Victoria Spooner and Trisha Starostina and Florian Strub and Yixuan Su and Jimin Sun and Dwarak Talupuru and Eugene Tarassov and Elena Tommasone and Jennifer Tracey and Billy Trend and Evren Tumer and Ahmet Üstün and Bharat Venkitesh and David Venuto and Pat Verga and Maxime Voisin and Alex Wang and Donglu Wang and Shijian Wang and Edmond Wen and Naomi White and Jesse Willman and Marysia Winkels and Chen Xia and Jessica Xie and Minjie Xu and Bowen Yang and Tan Yi-Chern and Ivan Zhang and Zhenyu Zhao and Zhoujie Zhao},
      year={2025},
      eprint={2504.00698},
      archivePrefix={arXiv},
      primaryClass={cs.CL},
      url={https://arxiv.org/abs/2504.00698}, 
}

@inproceedings{lv-etal-2024-full,
    title = "Full Parameter Fine-tuning for Large Language Models with Limited Resources",
    author = "Lv, Kai  and
      Yang, Yuqing  and
      Liu, Tengxiao  and
      Guo, Qipeng  and
      Qiu, Xipeng",
    editor = "Ku, Lun-Wei  and
      Martins, Andre  and
      Srikumar, Vivek",
    booktitle = "Proceedings of the 62nd Annual Meeting of the Association for Computational Linguistics (Volume 1: Long Papers)",
    month = aug,
    year = "2024",
    address = "Bangkok, Thailand",
    publisher = "Association for Computational Linguistics",
    url = "https://aclanthology.org/2024.acl-long.445/",
    doi = "10.18653/v1/2024.acl-long.445",
    pages = "8187--8198"
}

@inproceedings{qlora,
 author = {Dettmers, Tim and Pagnoni, Artidoro and Holtzman, Ari and Zettlemoyer, Luke},
 booktitle = {Advances in Neural Information Processing Systems},
 editor = {A. Oh and T. Naumann and A. Globerson and K. Saenko and M. Hardt and S. Levine},
 pages = {10088--10115},
 publisher = {Curran Associates, Inc.},
 title = {QLoRA: Efficient Finetuning of Quantized LLMs},
 url = {https://proceedings.neurips.cc/paper_files/paper/2023/file/1feb87871436031bdc0f2beaa62a049b-Paper-Conference.pdf},
 volume = {36},
 year = {2023}
}

@misc{le2012buildinghighlevelfeaturesusing,
      title={Building high-level features using large scale unsupervised learning}, 
      author={Quoc V. Le and Marc'Aurelio Ranzato and Rajat Monga and Matthieu Devin and Kai Chen and Greg S. Corrado and Jeff Dean and Andrew Y. Ng},
      year={2012},
      eprint={1112.6209},
      archivePrefix={arXiv},
      primaryClass={cs.LG},
      url={https://arxiv.org/abs/1112.6209}, 
}

@misc{wei2022finetunedlanguagemodelszeroshot,
      title={Finetuned Language Models Are Zero-Shot Learners}, 
      author={Jason Wei and Maarten Bosma and Vincent Y. Zhao and Kelvin Guu and Adams Wei Yu and Brian Lester and Nan Du and Andrew M. Dai and Quoc V. Le},
      year={2022},
      eprint={2109.01652},
      archivePrefix={arXiv},
      primaryClass={cs.CL},
      url={https://arxiv.org/abs/2109.01652}, 
}

@inproceedings{wang-etal-2022-super,
    title = "Super-{N}atural{I}nstructions: Generalization via Declarative Instructions on 1600+ {NLP} Tasks",
    author = "Wang, Yizhong  and
      Mishra, Swaroop  and
      Alipoormolabashi, Pegah  and
      Kordi, Yeganeh  and
      Mirzaei, Amirreza  and
      Naik, Atharva  and
      Ashok, Arjun  and
      Dhanasekaran, Arut Selvan  and
      Arunkumar, Anjana  and
      Stap, David  and
      Pathak, Eshaan  and
      Karamanolakis, Giannis  and
      Lai, Haizhi  and
      Purohit, Ishan  and
      Mondal, Ishani  and
      Anderson, Jacob  and
      Kuznia, Kirby  and
      Doshi, Krima  and
      Pal, Kuntal Kumar  and
      Patel, Maitreya  and
      Moradshahi, Mehrad  and
      Parmar, Mihir  and
      Purohit, Mirali  and
      Varshney, Neeraj  and
      Kaza, Phani Rohitha  and
      Verma, Pulkit  and
      Puri, Ravsehaj Singh  and
      Karia, Rushang  and
      Doshi, Savan  and
      Sampat, Shailaja Keyur  and
      Mishra, Siddhartha  and
      Reddy A, Sujan  and
      Patro, Sumanta  and
      Dixit, Tanay  and
      Shen, Xudong",
    editor = "Goldberg, Yoav  and
      Kozareva, Zornitsa  and
      Zhang, Yue",
    booktitle = "Proceedings of the 2022 Conference on Empirical Methods in Natural Language Processing",
    month = dec,
    year = "2022",
    address = "Abu Dhabi, United Arab Emirates",
    publisher = "Association for Computational Linguistics",
    url = "https://aclanthology.org/2022.emnlp-main.340/",
    doi = "10.18653/v1/2022.emnlp-main.340",
    pages = "5085--5109"
}
\bibliographystyle{icml2026}  %

\newpage
\appendix
\onecolumn

\section*{Appendix}
\section{Model Choice}
\label{app:model_choice}

\begin{table}[htbp]                                                     
\centering                                                                                  
\caption{Top 10 open-source models under 9B parameters on the LM Arena text leaderboard (February 2026).}                                                                     
\setlength{\tabcolsep}{3pt}
\begin{tabular}{lrc}                                        
\toprule
\textbf{Model} & \textbf{Score} & \textbf{Params} \\             
\midrule
Gemma 3n E4B IT          & 1319 & 4B \\
Gemma 3 4B IT            & 1303 & 4B \\
Ministral 8B             & 1237 & 8B \\
Llama 3 8B Instruct      & 1224 & 8B \\
Llama 3.1 8B Instruct    & 1212 & 8B \\
Gemma 2 2B IT            & 1199 & 2B \\
Gemma 1.1 7B IT          & 1180 & 7B \\
Phi-3 Small 8K           & 1172 & 7B \\
Llama 3.2 3B Instruct    & 1167 & 3B \\
Mistral 7B Instruct v0.2 & 1150 & 7B \\
\bottomrule
\end{tabular}
\label{tab:arena_small_models}
\end{table}

\section{Training Hyperparameters}
\label{app:hyperparams}

\begin{table}[htbp]
\centering
\caption{Training hyperparameters for LoRA and full fine-tuning on Mistral-7B-Instruct-v0.2 and Gemma models.}
\label{tab:hyperparameters}
\begin{tabular}{lcc}
\toprule
\textbf{Hyperparameter} & \textbf{LoRA} & \textbf{Full FT} \\
\midrule
Optimizer & AdamW & AdamW \\
Weight Decay & 0.01 & 0.01 \\
LR Schedule & Cosine & Cosine \\
LR Warmup & 10\% & 10\% \\
Batch Size & 32 & 32 \\
Epochs & 1 & 1 \\
\midrule
\multicolumn{3}{l}{\textit{LoRA-specific}} \\
\midrule
Rank ($r$) & 8 & -- \\
Alpha ($\alpha$) & 16 & -- \\
Target modules & qkv\_proj, MLP & -- \\
\bottomrule
\end{tabular}
\end{table}

\begin{table}[H]
    \centering
    \caption{\textbf{Selected learning rates.} Best learning rate per model, setup, method, and task, chosen by maximizing mean validation accuracy across seeds. LoRA rates are swept over $[5\mathrm{e}{-5},\, 5\mathrm{e}{-4}]$; full fine-tuning rates over $[5\mathrm{e}{-6},\, 5\mathrm{e}{-5}]$.}
    \label{tab:selected_lrs}
    \resizebox{\linewidth}{!}{
    \begin{tabular}{lllcccccccc}
        \toprule
        \textbf{Model} & \textbf{Setup} & \textbf{Method} & \textbf{ARC-e} & \textbf{CSQA} & \textbf{Hella.} & \textbf{MathQA} & \textbf{OBQA} & \textbf{PIQA} & \textbf{SIQA} & \textbf{Wino.} \\
        \midrule
        \multirow{4}{*}{Gemma-3 1B} & \multirow{2}{*}{LoRA} & From scratch & $2.7\text{e-}4$ & $5\text{e-}4$ & $2.7\text{e-}4$ & $2.7\text{e-}4$ & $5\text{e-}4$ & $3.8\text{e-}4$ & $3.8\text{e-}4$ & $3.8\text{e-}4$ \\
        & & Mashup Learning & $2.7\text{e-}4$ & $3.8\text{e-}4$ & $3.8\text{e-}4$ & $2.7\text{e-}4$ & $3.8\text{e-}4$ & $1.6\text{e-}4$ & $5\text{e-}4$ & $3.8\text{e-}4$ \\
        \cmidrule{2-11}
         & \multirow{2}{*}{Full FT} & From scratch & $1.6\text{e-}5$ & $2.7\text{e-}5$ & $2.7\text{e-}5$ & $1.6\text{e-}5$ & $2.7\text{e-}5$ & $2.7\text{e-}5$ & $2.7\text{e-}5$ & $2.7\text{e-}5$ \\
        & & Mashup Learning & $2.7\text{e-}5$ & $2.7\text{e-}5$ & $2.7\text{e-}5$ & $1.6\text{e-}5$ & $2.7\text{e-}5$ & $1.6\text{e-}5$ & $1.6\text{e-}5$ & $2.7\text{e-}5$ \\
        \midrule
        \multirow{4}{*}{Gemma-2 2B} & \multirow{2}{*}{LoRA} & From scratch & $2.7\text{e-}4$ & $3.8\text{e-}4$ & $1.6\text{e-}4$ & $1.6\text{e-}4$ & $3.8\text{e-}4$ & $2.7\text{e-}4$ & $1.6\text{e-}4$ & $1.6\text{e-}4$ \\
        & & Mashup Learning & $3.8\text{e-}4$ & $2.7\text{e-}4$ & $1.6\text{e-}4$ & $2.7\text{e-}4$ & $3.8\text{e-}4$ & $3.8\text{e-}4$ & $1.6\text{e-}4$ & $1.6\text{e-}4$ \\
        \cmidrule{2-11}
         & \multirow{2}{*}{Full FT} & From scratch & $5\text{e-}6$ & $2.7\text{e-}5$ & $5\text{e-}6$ & $5\text{e-}6$ & $2.7\text{e-}5$ & $5\text{e-}6$ & $5\text{e-}6$ & $5\text{e-}6$ \\
        & & Mashup Learning & $5\text{e-}6$ & $2.7\text{e-}5$ & $5\text{e-}6$ & $5\text{e-}6$ & $2.7\text{e-}5$ & $5\text{e-}6$ & $5\text{e-}6$ & $5\text{e-}6$ \\
        \midrule
        \multirow{4}{*}{Gemma-3 4B} & \multirow{2}{*}{LoRA} & From scratch & $3.8\text{e-}4$ & $5\text{e-}4$ & $2.7\text{e-}4$ & $3.8\text{e-}4$ & $5\text{e-}4$ & $5\text{e-}4$ & $3.8\text{e-}4$ & $2.7\text{e-}4$ \\
        & & Mashup Learning & $5\text{e-}4$ & $3.8\text{e-}4$ & $2.7\text{e-}4$ & $2.7\text{e-}4$ & $5\text{e-}4$ & $2.7\text{e-}4$ & $1.6\text{e-}4$ & $2.7\text{e-}4$ \\
        \cmidrule{2-11}
         & \multirow{2}{*}{Full FT} & From scratch & $5\text{e-}6$ & $2.7\text{e-}5$ & $2.7\text{e-}5$ & $2.7\text{e-}5$ & $2.7\text{e-}5$ & $2.7\text{e-}5$ & $5\text{e-}6$ & $2.7\text{e-}5$ \\
        & & Mashup Learning & $5\text{e-}6$ & $2.7\text{e-}5$ & $2.7\text{e-}5$ & $2.7\text{e-}5$ & $2.7\text{e-}5$ & $2.7\text{e-}5$ & $5\text{e-}6$ & $2.7\text{e-}5$ \\
        \bottomrule
    \end{tabular}
    }
\end{table}

\section{Per-Task Learning Rate Sensitivity}
\label{app:lr_sensitivity_tasks}

\begin{figure}[H]
    \centering
    \begin{subfigure}[t]{\columnwidth}
        \centering
        \includegraphics[width=\linewidth]{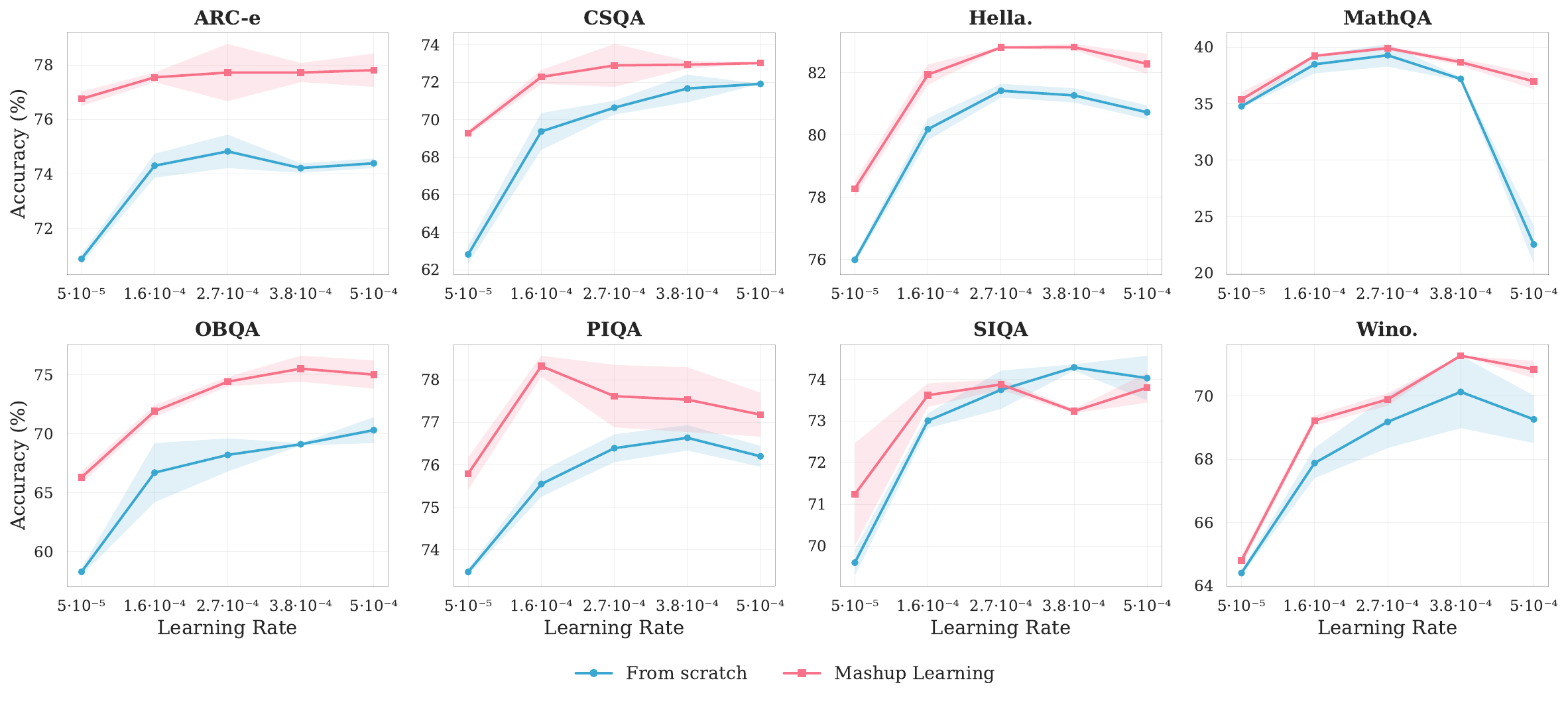}
        \caption{LoRA (3 seeds)}
    \end{subfigure}
    \begin{subfigure}[t]{\columnwidth}
        \centering
        \includegraphics[width=\linewidth]{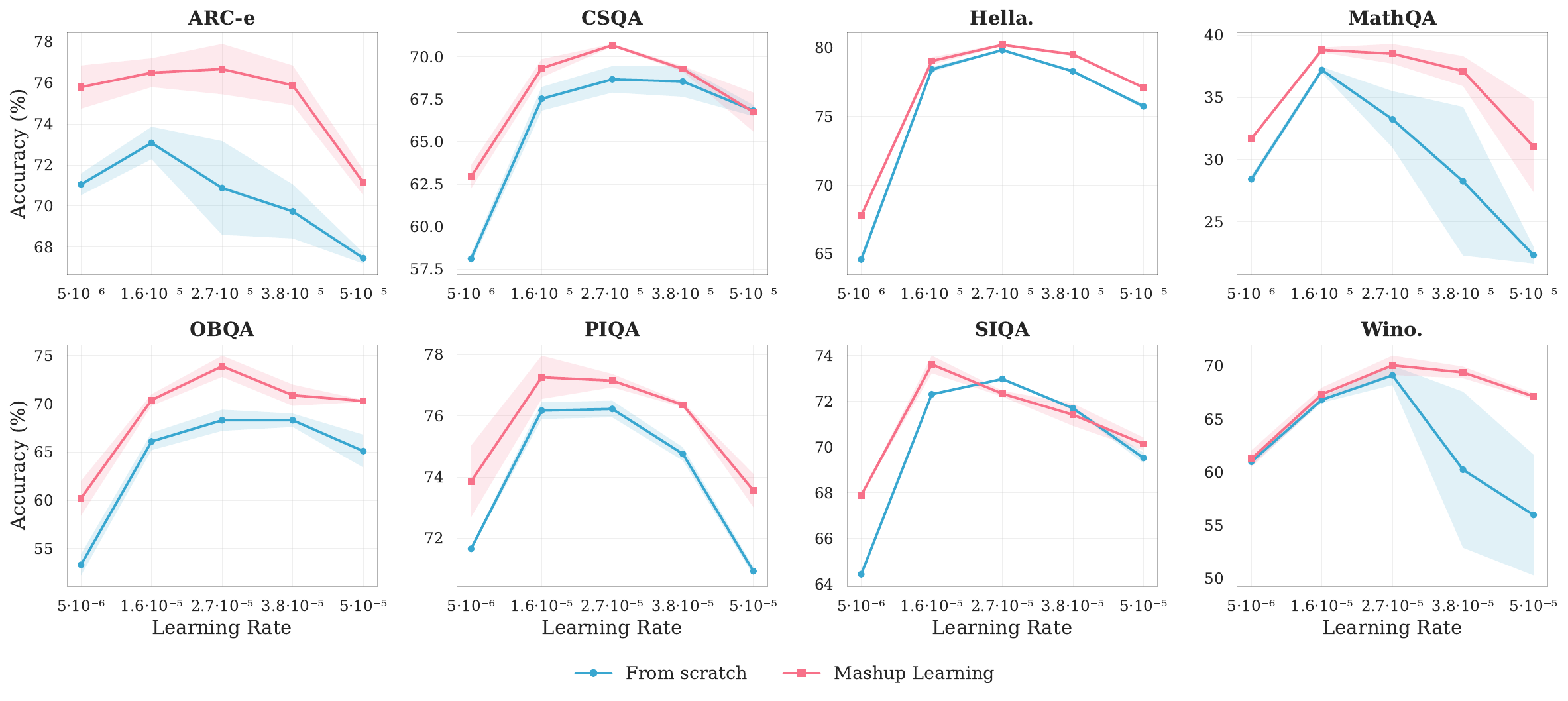}
        \caption{Full FT (3 seeds)}
    \end{subfigure}
    \caption{Gemma-3 1B: per-task learning rate sensitivity.}
    \vspace{12em}
    \label{fig:gemma3_1b_lr_sensitivity_tasks}
\end{figure}

\begin{figure}[H]
    \centering
    \begin{subfigure}[t]{\columnwidth}
        \centering
        \includegraphics[width=\linewidth]{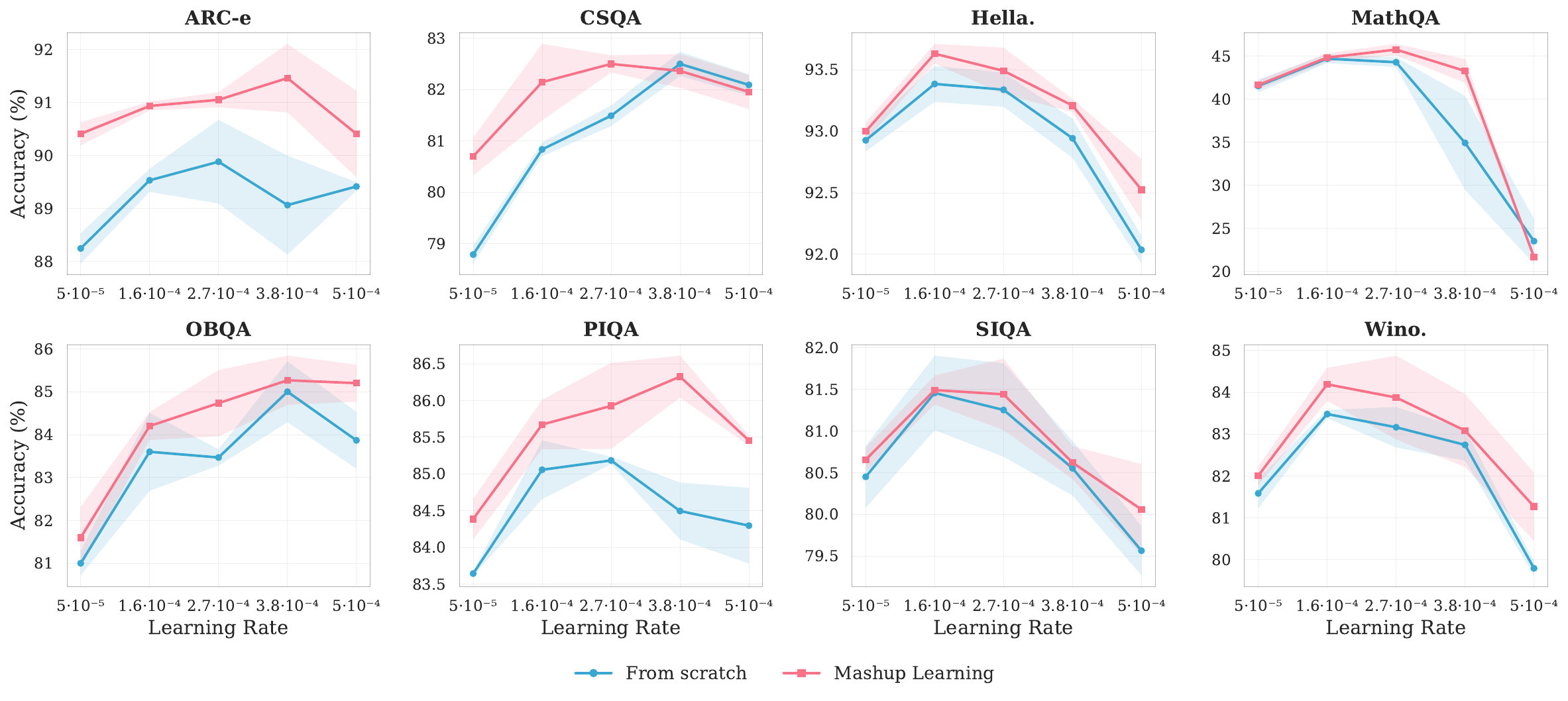}
        \caption{LoRA (3 seeds)}
    \end{subfigure}
    \begin{subfigure}[t]{\columnwidth}
        \centering
        \includegraphics[width=\linewidth]{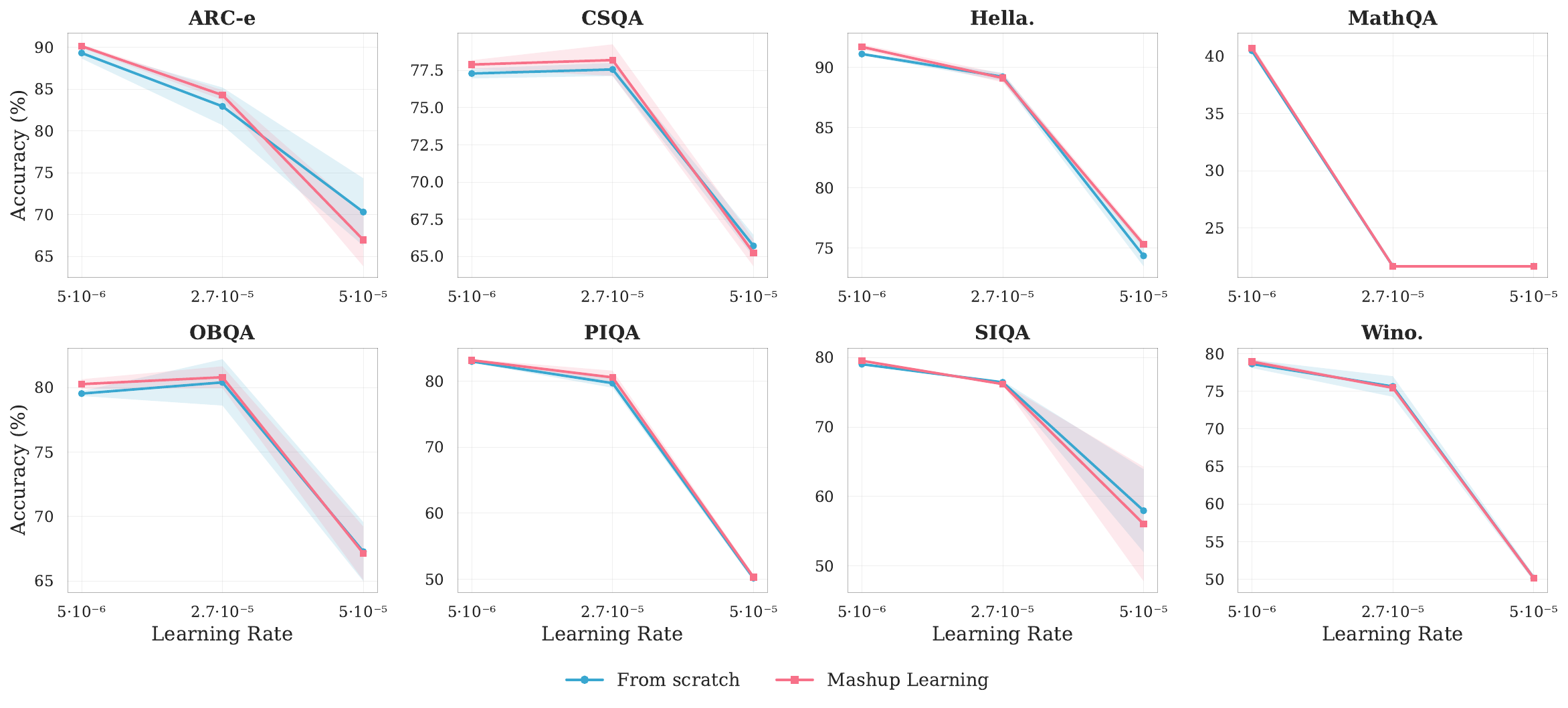}
        \caption{Full FT (3 seeds)}
    \end{subfigure}
    \caption{Gemma-2 2B: per-task learning rate sensitivity.}
    \label{fig:gemma2_2b_lr_sensitivity_tasks}
\end{figure}

\begin{figure}[H]
    \centering
    \begin{subfigure}[t]{\columnwidth}
        \centering
        \includegraphics[width=\linewidth]{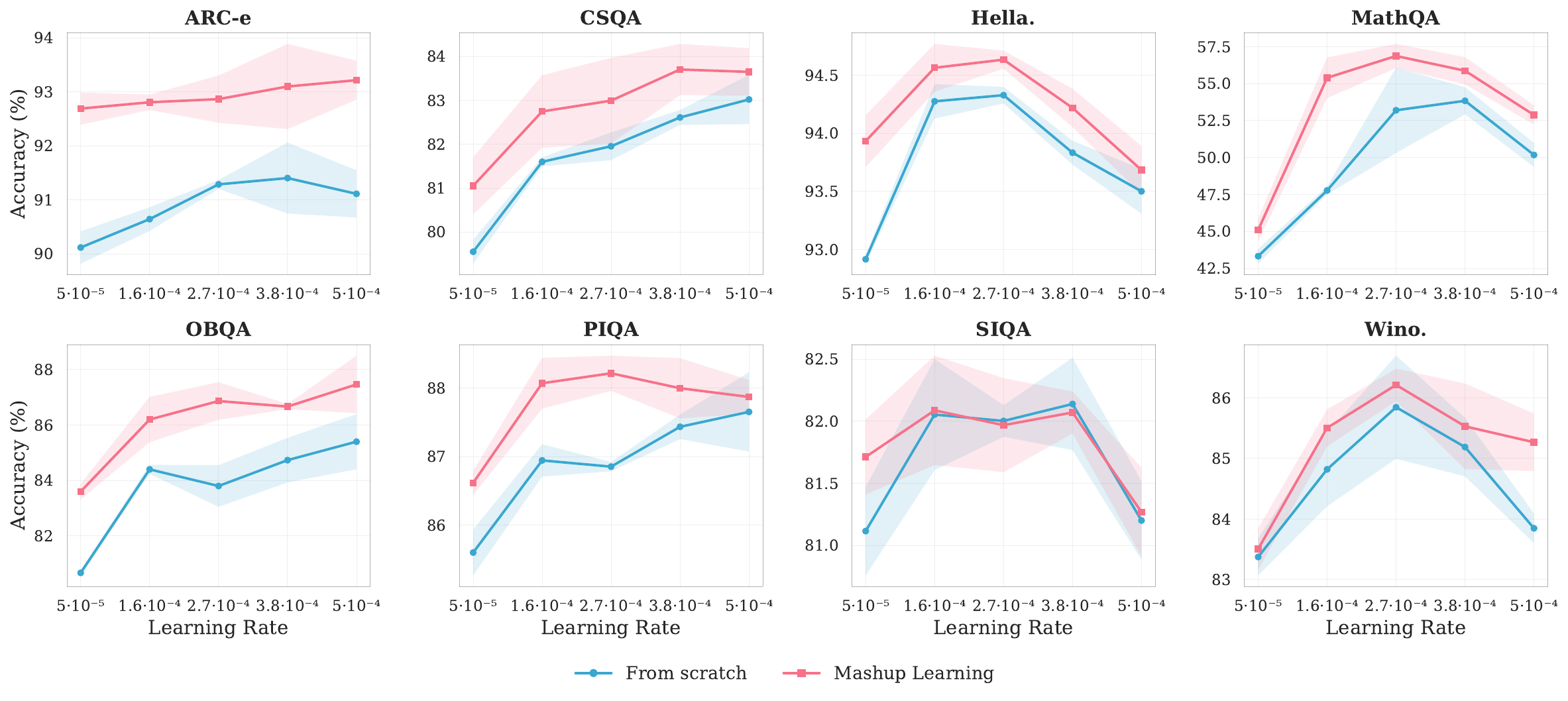}
        \caption{LoRA (3 seeds)}
    \end{subfigure}
    \begin{subfigure}[t]{\columnwidth}
        \centering
        \includegraphics[width=\linewidth]{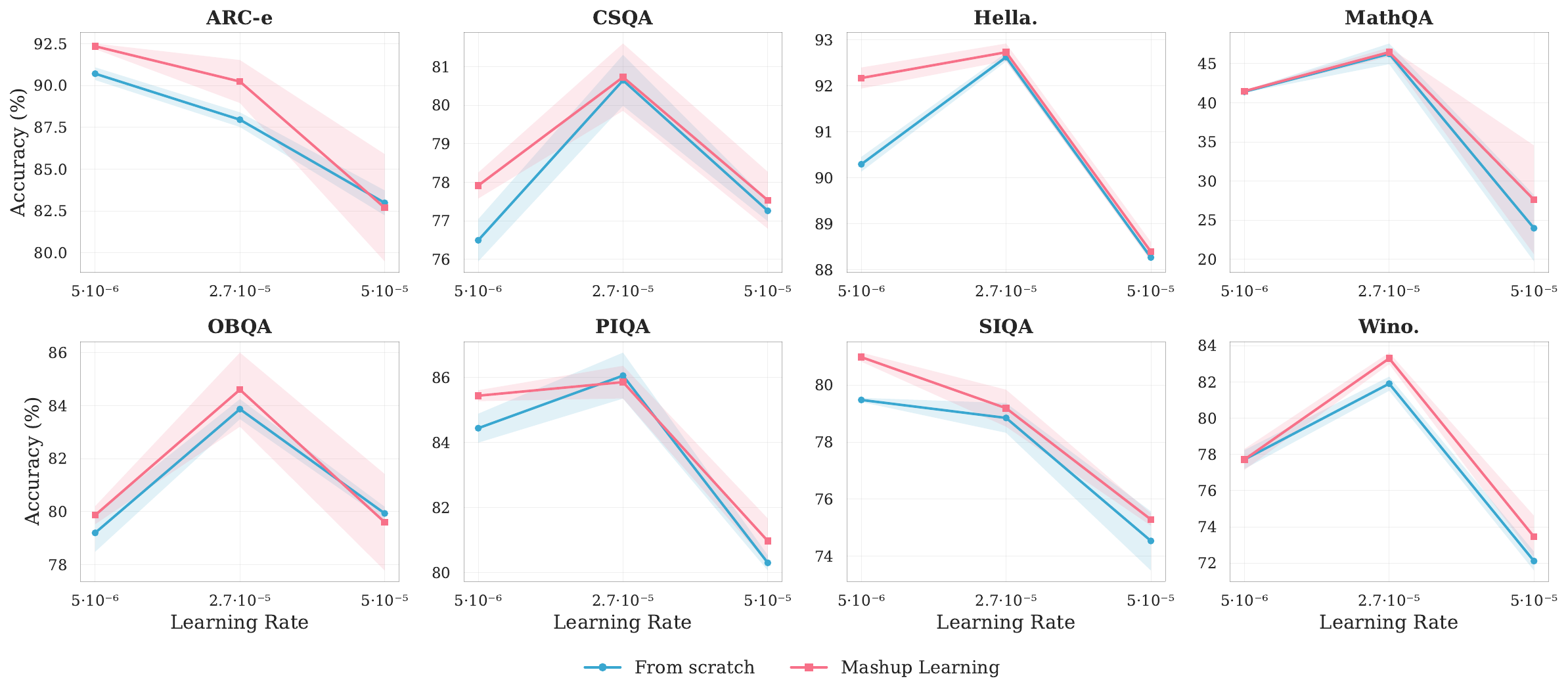}
        \caption{Full FT (3 seeds)}
    \end{subfigure}
    \caption{Gemma-3 4B: per-task learning rate sensitivity.}
    \label{fig:gemma3_4b_lr_sensitivity_tasks}
\end{figure}

\end{document}